%% file: main.tex
\documentclass[10pt,twocolumn,letterpaper]{article}
\input{math_commands.tex}

\usepackage{cvpr}              %

\input{preamble}

\definecolor{cvprblue}{rgb}{0.21,0.49,0.74}
\usepackage[pagebackref,breaklinks,colorlinks,citecolor=cvprblue]{hyperref}

\usepackage{float}
\usepackage[dvipsnames]{xcolor}

\def\paperID{8816} %
\def\confName{CVPR}
\def\confYear{2024}
\def\methodFull{Fairy}

\title{\methodFull: \underline{Fa}st Parallellized \underline{I}nst\underline{r}uction-Guided Video-to-Video S\underline{y}nthesis}

\author{Bichen Wu \quad Ching-Yao Chuang \quad Xiaoyan Wang \quad Yichen Jia \\
Kapil Krishnakumar \quad Tong Xiao \quad Feng Liang \quad Licheng Yu \quad Peter Vajda \\
GenAI, Meta \\
Project page: \url{https://fairy-video2video.github.io}
}

\definecolor{ForestGreen}{RGB}{34,139,34}

\def\ourmodel{Fairy}   %
\def\ourmodell{Fairy } %

\begin{document}

\twocolumn[{
\renewcommand\twocolumn[1][]{#1}
\maketitle
\centering
\vspace*{-3mm}
\includegraphics[width=1\textwidth]{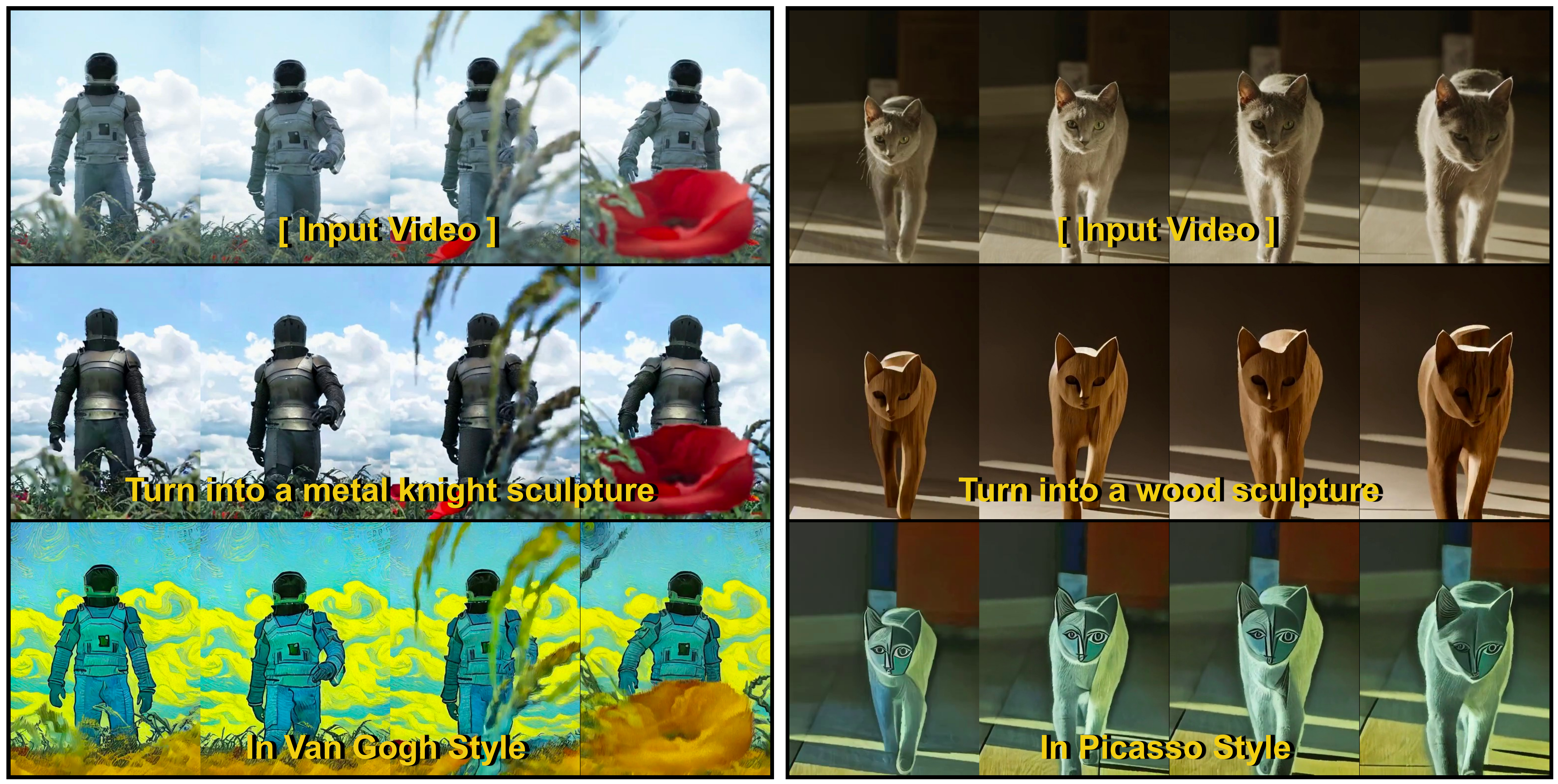}
\vspace*{-6mm}
\captionof{figure}{\textbf{\ourmodell for Instruction-Guided Video Editing.} Given \textbf{a video} and an \textbf{instruction} for editing, \ourmodell performs accurate edits while ensuring temporal coherence. Remarkably efficient, 120 frames of 512$\times$384 video can be generated in just \textbf{14 seconds}. We refer readers to our supplementary material to check the results in video format.
\vspace*{8mm}
}\label{fig:teaser}
}]

\maketitle

\input{sec/0_abstract}    
\input{sec/1_intro}

\input{sec/2_related_work}
\input{sec/3_preliminaries}

\input{sec/4_crossframe_att}

\input{sec/5_method}

\input{sec/6_experiments}

\input{sec/7_conclusion}
\clearpage
{
    \small
    \bibliographystyle{ieeenat_fullname}
    \bibliography{main}
}

\appendix
\input{sec/appendix}

\end{document}

%% file: math_commands.tex
\usepackage{amsmath,amsfonts,bm}

\def\eqref#1{equation~(\ref{#1})}

\def\1{\bm{1}}

\def\cT{{\mathcal{T}}}

\def\mA{{\bm{A}}}

\def\mI{{\bm{I}}}

\def\mK{{\bm{K}}}

\def\mQ{{\bm{Q}}}

\def\mV{{\bm{V}}}

\DeclareMathAlphabet{\mathsfit}{\encodingdefault}{\sfdefault}{m}{sl}
\SetMathAlphabet{\mathsfit}{bold}{\encodingdefault}{\sfdefault}{bx}{n}

\def\gI{{\mathcal{I}}}

\def\gT{{\mathcal{T}}}

\def\sR{{\mathbb{R}}}

\newcommand{\softmax}{\mathrm{softmax}}

\DeclareMathOperator*{\argmax}{arg\,max}

%% file: preamble.tex
\usepackage[dvipsnames]{xcolor}

%% file: sec/0_abstract.tex
\begin{abstract}
In this paper, we introduce \ourmodel, a minimalist yet robust adaptation of image-editing diffusion models, enhancing them for video editing applications. Our approach centers on the concept of anchor-based cross-frame attention, a mechanism that implicitly propagates diffusion features across frames, ensuring superior temporal coherence and high-fidelity synthesis. \ourmodell not only addresses limitations of previous models, including memory and processing speed. It also improves temporal consistency through a unique data augmentation strategy. This strategy renders the model equivariant to affine transformations in both source and target images. Remarkably efficient, \ourmodell generates 120-frame 512$\times$384 videos (4-second duration at 30 FPS) in just 14 seconds, outpacing prior works by at least 44$\times$. A comprehensive user study, involving 1000 generated samples, confirms that our approach delivers superior quality, decisively outperforming established methods.

\end{abstract}

%% file: sec/1_intro.tex
\section{Introduction}
\label{sec:intro}

The advent of generative artificial intelligence has heralded a new era of creative potential, characterized by the ability to create or modify content in an effortless manner. In particular, image editing has undergone significant evolution, driven by text-to-image diffusion models pretrained on billion-scale datasets. This surge has catalyzed a vast array of applications in image editing and content creation.

Building on the accomplishments of image-based models, the next natural frontier is transitioning these capabilities to the temporal dimension to enable effortless and creative video editing. A direct strategy to leap from image to video, is to simply process a video on a frame-by-frame basis using an image model. Nonetheless, generative image editing is inherently high-variance -- there are countless ways to edit a given image based on the same text prompt. As a result, it is difficult to maintain temporal coherence if each frame is edited independently \citep{wang2023zero}.

Previous and concurrent studies have proposed several ways to improve the temporal consistency, and one promising paradigm is what we call \textit{tracking-and-propagation}: one first apply an image editing model on one or a few frames, then tracks pixels across all frames and propagate the edit to the entire video. Existing works \cite{yang2023rerender,warpfusion,jamrivska2019stylizing,bar2022text2live,chai2023stablevideo,kasten2021layered,lee2023shape,ouyang2023codef} track pixels mainly through optical flow or by reconstructing videos as some canonical layered representations. Despite some successful applications, this paradigm is not robust, since tracking is an unsolved computer vision challenge. Existing methods, including optical flow or layered video representation, often struggle with videos with large motion and complex dynamics.

In this work, we introduce \emph{\ourmodel}, a versatile and efficient video-to-video synthesis framework that generates high-quality videos with remarkable speed (Figure \ref{fig:teaser}). Our work re-examines the tracking-and-propagation paradigm under the context of diffusion model features. In particular, we bridge cross-frame attention with correspondence estimation,
showing that it  temporally tracks and propagates intermediate features inside a diffusion model. %
The cross-frame attention map can be interpreted as a similarity metric assessing the correspondence between tokens throughout various frames, where features from one semantic region will assign higher attention to similar semantic regions in other frames, as shown in Figure \ref{fig_tracking_example}.
Consequently, the current feature representations are refined and propagated through a weighted sum of similar regions across frames via attention, effectively minimizing feature disparity between frames, which translates to improved temporal consistency.

The analysis gives rise to our \emph{anchor-based model}, the central component of \ourmodel. To ensure temporal consistency, we sample $K$ anchor frames from which we extract diffusion features, and the extracted features define a set global features to be propagated to successive frames.
When generating each new frame, we replace the self-attention layer with cross-frame attention with respect to the cached features of anchor frames. With cross-frame attention, the tokens in each frame take the features in anchor frames that exhibit analogous semantic content, thereby enhancing consistency. In addition, by sampling $K$ anchor frames instead of computing cross-attention with respect to all frames, \ourmodell 
achieves several advantages: (1) it ensures temporal consistency by sharing the same global features, (2) it overcomes the memory issue due to extensive frame number, (3) it enhances processing speed through the caching of anchor frame features, and (4) it streamlines parallel computation, thereby facilitating remarkably rapid generation on multiple GPUs.

Despite the improvement from anchor-based cross-frame attention, the model is still sensitive to minor variations within the input frames, even with the same text prompt and initial latent noise. Such small changes could stem from natural movements within a video sequence or from affine transformations applied to the input. The gold standard solution is to train the model with pairs of original and edited videos, thereby accommodate it to recognize and adapt to such variations. However, collecting such a dataset is far from straightforward. To emulate these transformations, we employ a data augmentation strategy. Starting with an input image and its edited counterpart, we apply a sequence of affine transformations to both, generating successive frames. The assumption is that the affine transformations applied to input images should correspondingly affect the edited images. This method of \emph{equivariant finetuning} leads to notable enhancements in temporal consistency.

To verify the effectiveness of \ourmodel, we conduct a large-scale evaluation consists of 1000 generated videos. Both human evaluation and quantitative metrics confirm that our model achieves significantly better quality. Moreover, thanks to the simplicity of the design and the parallelizable architecture, \ourmodell achieves $>$44x speedup over baselines.  

In short, this work makes the following contributions: 
(1) We adopt a series of simple yet effective adaptions that transform an image-editing model for video-to-video synthesis. 
(2) We evaluate our approach via extensive human study with 1000 generated videos, confirming that \ourmodell delivers superior quality over prior state-of-the art methods.
(3) \ourmodel~ is blazing fast, achieving $>$44x speedup over previous methods when utilizing 8-gpu parallel generation.

%% file: sec/2_related_work.tex
\section{Related works}
\label{sec:related_works}
\textbf{Conditional video generation}: 
Following the success of diffusion models in text-to-image generation \cite{saharia2022photorealistic,ramesh2022hierarchical,dai2023emu,rombach2022high}, there has been a surge in video generation. Based on a text-to-video model, video-to-video generation can be achieved by conditioning the model on attributes extracted from a source video. For example, Gen-1 \cite{esser2023structure} conditions on the estimated depth while VideoComposer \cite{wang2023videocomposer} integrates additional cues, such as depth, motion vectors, sketches, among others. Building such models requires training on video datasets, which are much more scarce than image datasets \cite{schuhmann2022laion}. Training such models also imposes considerable computational demands. Consequently, these constraints confine video models to reduced resolution, shorter duration, and smaller model size, leading to a decline in visual quality when contrasted with contemporary image generation models. In comparison, our model is adapted from a pretrained image-to-image model. Our finetuning only requires image data, and the training cost (30 hours on 8 A100 GPUs) is substantially smaller than video models.

\textbf{Tracking and propagation}: this paradigm involves initiating edits on a single image, identifying pixel correspondences across the video sequence, then propagating the edit.
The key in this approach lies in tracking. Numerous efforts \cite{yang2023rerender,warpfusion,jamrivska2019stylizing} have adopted optical flow, keypoint tracking, or other motion cues to tackle this. Another stream of efforts \cite{ouyang2023codef, bar2022text2live, chai2023stablevideo, kasten2021layered, lee2023shape} reconstruct the video using a multi-layer canonical representation, associating pixels to canonical points on the representation. However, video tracking is an unsolved computer vision challenge and often fails on complex videos. Additionally, tracking-and-propagation does not allow edits to alter object contours, which breaks the pixel correspondence. Instead of tracking in pixel space, our model leverages cross-frame attention to implicitly track corresponding regions and propagate features to reduce frame discrepancy. Owing to the robustness and versatility of diffusion features, as also observed in \citet{tang2023emergent}, our approach accommodates a broader spectrum of videos and offers enhanced editing flexibility.

\textbf{Image model adaptation}: Many works also adapt image-to-image models to video. For example, \cite{khachatryan2023text2video} modifies self-attention in diffusion models. \cite{wu2023tune} performs per-video finetuning and utilizes a inversion-denoising procedure for editing. \cite{vid2vid-zero,liu2023video,qi2023fatezero,geyer2023tokenflow} adapt image-to-image pipelines \cite{hertz2022prompt, tumanyan2023plug, brooks2023instructpix2pix} to edit videos, by modifying/adding cross-frame attention modules, null-text inversion, etc. Most of these methods can only generate video clips with a small number of frames, while \cite{geyer2023tokenflow} leverages a nearest-neighbor field on diffusion features to propagate key frame features to more frames. Our model improves the design of spatial temporal attention \cite{khachatryan2023text2video,vid2vid-zero,liu2023video,qi2023fatezero} to anchor-based cross-frame attention, which enables generating long videos with arbitrarily many frames. We further improves its temporal consistency by equivariant finetuning. Our work bears resemblance to the concurrent work \cite{geyer2023tokenflow}. To edit a video, 
\cite{geyer2023tokenflow} first performs a latent inversion on the original video, extract a nearest-neighbor field, which is then used for feature propagation to generate the target video. Our pipeline is much simpler and more efficient. We do not require latent inversion; and the feature propagation is achieved through attention; our architectures naturally allows parallel generation. As a result, our model is 53 times faster than \cite{geyer2023tokenflow}.

%% file: sec/3_preliminaries.tex
\section{Preliminaries}

\paragraph{Video-to-Video Diffusion Models}
In this work, we primarily focus on instruction-guided video editing. Given an input video with $N$ frames $\mI = \{I^1, \dots, I^N\} \in \gI^N$, the goal is to edit it into a new video $\mI' = \{{I^1}', \dots, {I^N}'\} \in \gI^N$ according to an natural language instruction $c \in \cT$ that preserves the semantic of the original video. A straightforward baseline is to adopt an image-based editting model $f: (\gI, \gT) \rightarrow \gI$ to edit the video frame by frame: $\mI' = \{f(I^1, c), \dots, f(I^N, c)\}$. In this work, we build upon this line of work and improve the consistency with a variant of cross-frame attention.

\paragraph{Self-attention and Cross-frame attention}
Self-attention has played a crucial role in the diffusion networks. In a self-attention block, features of tokens are projected into
queries $\mQ \in \sR^{n \times d}$, keys $\mK \in \sR^{n \times d}$, and values $\mV \in \sR^{n \times d}$, where the output is defined as
\begin{align*}
    \mathrm{SelfAttention}(\mQ, \mK, \mV) = \softmax \left( \frac{\mQ \mK^T}{\sqrt{d}} \right) \mV.
\end{align*}
The output from the softmax is commonly referred to as the \emph{attention score} or \emph{attention map}. Given $N$ frames, to extend the self-attention to cross-frame attention, one can simply concat the keys and values from all frames, e.g., $\mK^\ast = [\mK^1, \cdots, \mK^N]$, and compute the self-attention as $\textnormal{Self-Attention}(\mQ, \mK^\ast, \mV^\ast)$. In particular, cross-frame attention provides temporal modeling capability by attending to other frames and shows encouraging results in improving temporal consistency \citep{wang2023zero, liu2023video}.

%% file: sec/4_crossframe_att.tex
\section{Implicit Tracking via Cross-frame Attention}
We first bridge cross-frame attention with correspondence estimation, fostering a straightforward yet effective feature propagation mechanism for video-to-video generation.

The primary objective of self-attention is to select appropriate values $\mV$ with the attention scores determined by $\mQ \mK^T$. In the case of cross-frame attention, given a token location $p$ in a frame, the attention score is computed by the cosine similarity between $\mQ_{p, :}$ and each token in $\mK^\ast$, where the key values $\mV^\ast$ are the features of tokens across both spatial and temporal dimension. 

It is noteworthy that the mathematical formulation exhibits profound similarities to feature propagation mechanisms. Specifically, the attention score serves as the estimated correspondence, and the output of attention module could be interpreted as a fused representation of warped features derived from successive frames. We will empirically substantiate this hypothesis through analyses of the tracking behavior inherent in the attention score. 

\begin{figure}[!tb]
\begin{center}   \includegraphics[width=\linewidth]{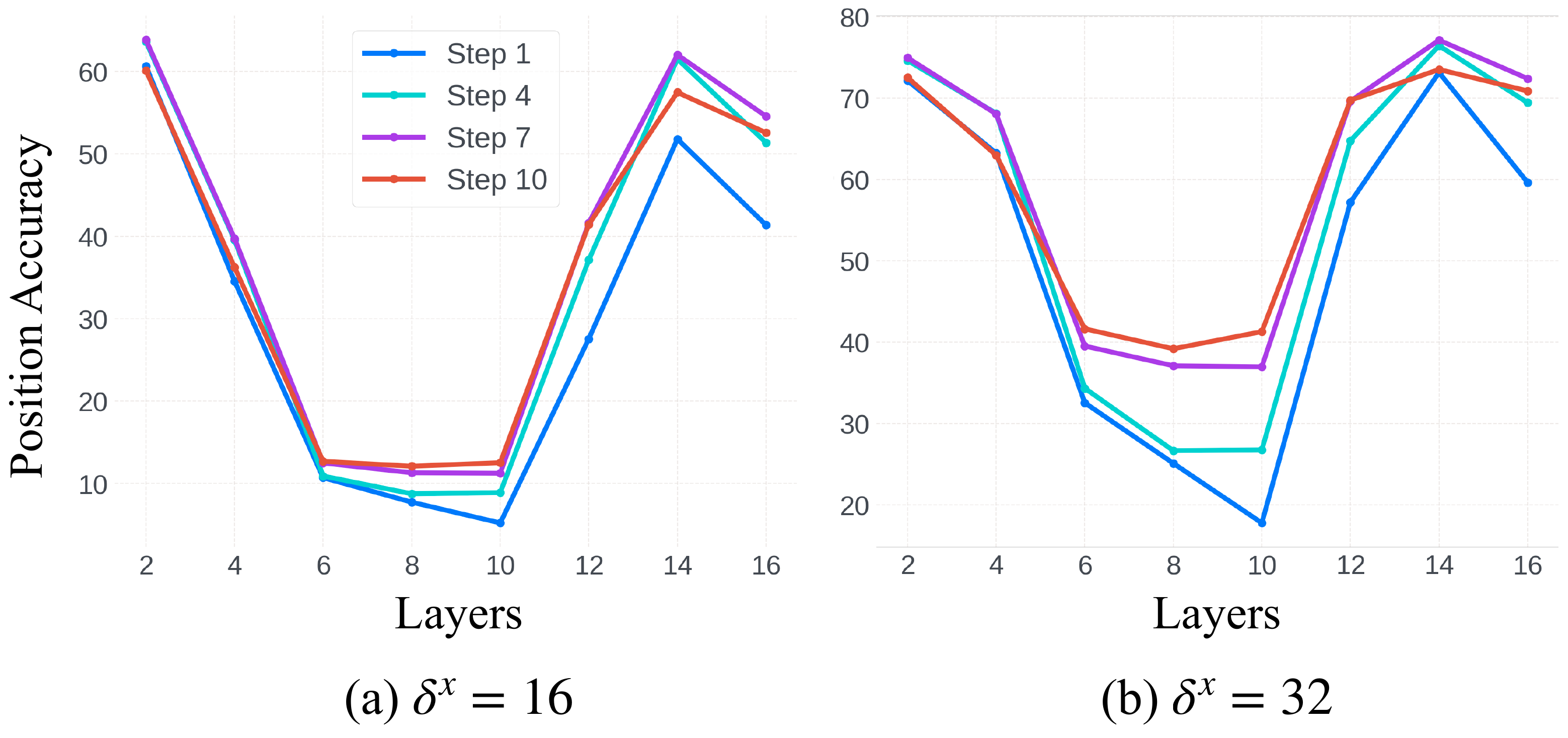}
\end{center}
\vspace{-3.5mm}
   \caption{\textbf{Position Accuracy $\delta^x$ on DAVIS.} The cross-frame attention score demonstrates significant tracking proficiency, particularly evident in the initial and final stages of the UNet. } 
\label{fig_tracking_tap}
\vspace{-2.5mm}
\end{figure}

\subsection{Temporal Tracking with Attention Score}
In this section, we provide evidences that the attention scores in cross-frame attention implicitly serve as a correspondence estimation across frames. In particular, we adopt a conditional image-to-image diffusion model and examine the attention map between two frames of a video clip. Consider $\mQ^t$ and $\mK^t$ as the respective query and key representations corresponding to the frame at time $t$. To corroborate our conjecture regarding the role of attention scores, we designate a specific query point $p$ at time $t$ and endeavor to ascertain its corresponding coordinate $q$ at a subsequent time $t'$ through the expression:
\begin{align*}
    q = \argmax_{p'} \mA_{p, p'}, \; \textnormal{where} \; \mA = \softmax(\frac{\mQ^t {\mK^{t'}}^T}{\sqrt{d}}),
\end{align*}
where $\mA_{p, p'}$ denotes the element of the matrix $\mA$ located at the row index $p$ and column index $p'$. The correspondence is estimated by selecting the location $p'$ with the highest attention score with respect to $p$. For multi-head attention, we average the attention scores from all heads. By evaluating the tracking ability of the proposed estimator, we can verify whether the attention scores are good correspondence estimator for feature propagation.

\subsection{Video Tracking Experiments: TAP-Vid}
In our evaluation, we utilize the DAVIS datasets from the TAP-Vid \citep{doersch2022tap, pont20172017}, with 30 videos clips ranging from 34-104 frames. The frames are resize to $256 \times 256$ for evaluation. We measure the $<\delta^x$ position accuracy proposed in TAP-Vid, which calculates the fraction of points that are within $\delta^x$ pixels of their ground truth position. The dimensions of the attention map inherently impose a constraint on the precision achievable in point tracking. Since diffusion UNets adopts spatial downsampling, 
we configure $\delta^x$ at the values of 16 and 32 for our experiments. We set the number of diffusion step to $10$ with Euler ancestral sampler \citep{karras2022elucidating}.

\begin{figure}[!tb]
\begin{center}   \includegraphics[width=\linewidth]{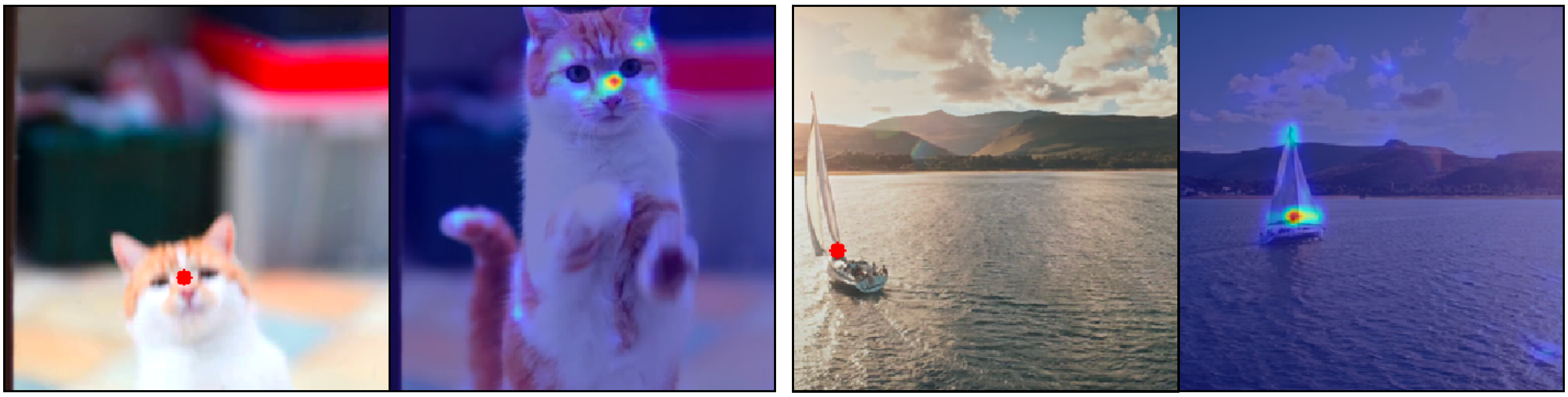}
\end{center}
\vspace{-3.5mm}
   \caption{\textbf{Visualization of Attention Score.} The left image shows the query point $p$ within the current frame, and the right image is the target frame. Cross-frame attention performs accurate temporal correspondence estimation without any finetuning. } 
\label{fig_tracking_example}
\vspace{-2.5mm}
\end{figure}

Figure \ref{fig_tracking_tap} shows the position accuracy for attention scores across different layers and diffusion step. We can see that the first and last few layers demonstrate a strong tracking results, achieving over $ 60 \% / 70 \%$ accuracy for $\delta^x = 16/32$. Interestingly, the results are consistent across different diffusion step, demonstrating the strong tracking ability of cross-frame attention. The observed degradation in accuracy at the central layers of the UNet architecture can be attributed primarily to the reduction in the spatial resolution of the feature maps. For instance, within the seventh layer of the network, the feature map dimensions are constrained to $4 \times 4$. Figure \ref{fig_tracking_example} visualizes the attention score on a target frame given a query point. We can see that the attention map locate the corresponding position in target frame.

\paragraph{Cross-frame Attention $\approx$ Tracking and Feature Propagation} Our experimental findings disclose an unexpectedly potent tracking capability associated with the attention score. These results robustly validate our hypothesis: \emph{even in the absence of explicit finetuning, cross-frame attention implicitly executes a formidable feature propagation mechanism}. In particular, features $\mV^\ast$ from alternative frames are transmitted to the current frame based on the correspondence determined through the attention scores.

%% file: sec/5_method.tex
\section{\ourmodel: Fast Video-to-Video Synthesis}

Building on the analyses, we present \ourmodel, an efficacious video-to-video framework that leverages the inherent feature propagation of cross-frame attention. In particular, we propose to propagate the value features from a collection of anchor frames to a candidate frame using cross-frame attention. The performance can be further enhanced through the proposed equivariant finetuning method. We also demonstrate that \ourmodel ~is easily parallelizable, facilitating fast generation of arbitrarily long videos.

\begin{figure}[!tb]
\begin{center}   \includegraphics[width=\linewidth]{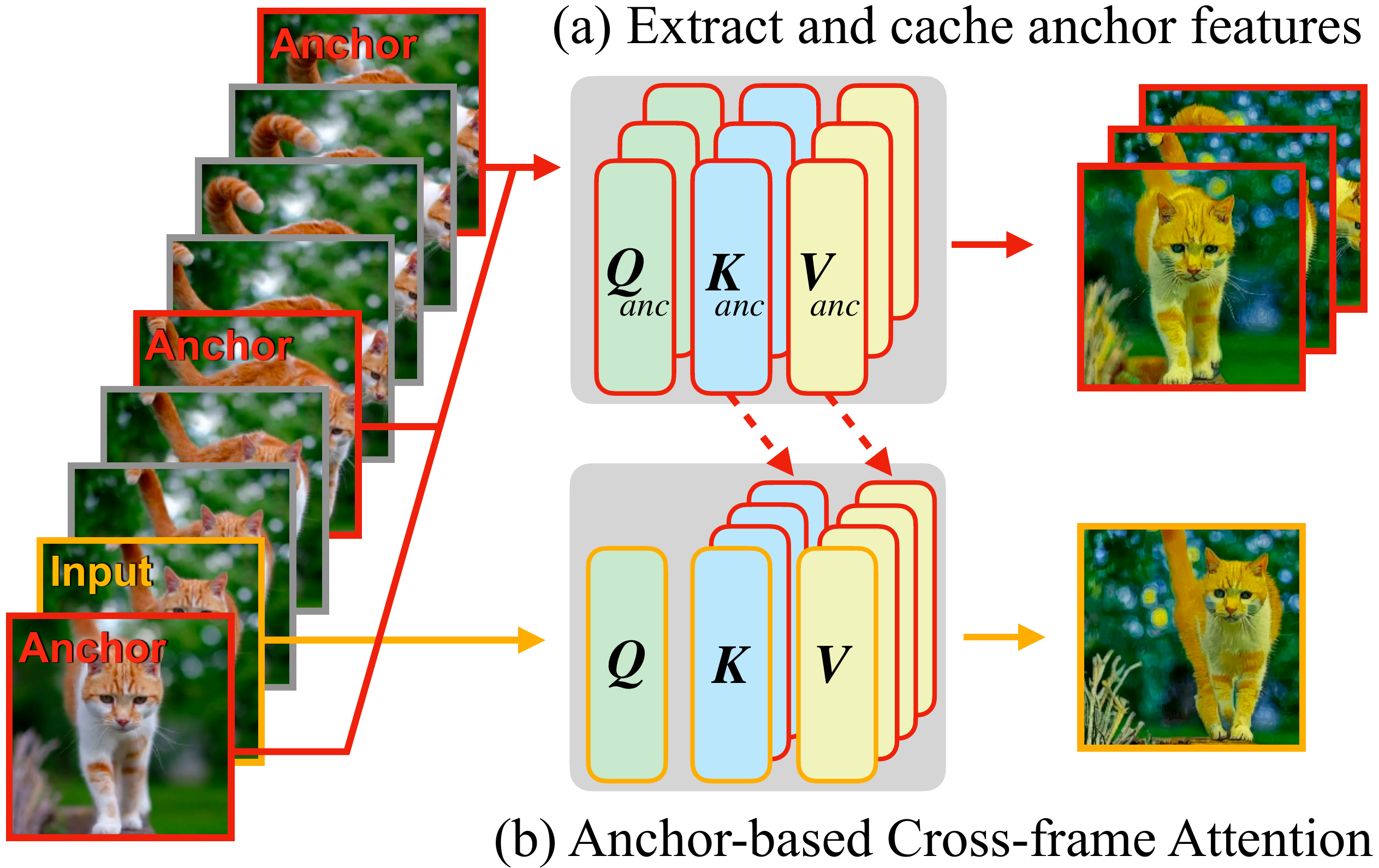}
\end{center}
\vspace{-3.5mm}
   \caption{\textbf{Illustration of Attention Blocks} (a) Given a set of anchor frames, we extract and cache the attention feature $\mK_{anc}$ and $\mV_{anc}$. (b) Given an input frame, we perform cross-frame attention with respect to the cached features of anchor frames.} 
\label{fig_cross_frame_attn}
\vspace{-2.5mm}
\end{figure}

\subsection{Anchor-Based Model}
Inspired by prior research in tracking-and-propagation, where the edits to one or a few frames are propagated to the entire video, we sample a set of \emph{anchor frames} and edit them with an image-based model $f: (\gI, \gT) \rightarrow \gI$. Similarly, our objective is to extend the edits in the anchor frames to the successive frames, but utilizing cross-frame attention mechanisms instead of optical flow or explicit point tracking. In particular, given a set of anchor frames $\mI_{anc} = \{\tilde{I}^1, \dots, \tilde{I}^K \} \subseteq \mI = \{ I^1, \dots, I^N \}$, we treat them as a batch and feed them to the diffusion model $f$, where the self-attention in the model is replaced with cross-frame attention in a zero-shot manner. Throughout the diffusion process, for each anchor frame $\tilde{I}^n$, we store its key and value vectors ${\mK}^{n, l, t}, {\mV}^{n, l, t}$ for every cross-frame attention layer $l$ and every diffusion step $t$ in cache. Intuitively, $\mV^{n, l, t}$ defines a set global features to be propagated to successive frames. To simplify the notation, we will drop the subscript $l$ and $t$ in the following sections.

Let $\mK_{anc} = [\mK^1, \dots, \mK^K]$ and $\mV_{anc} = [\mV^1, \dots, \mV^K]$ be the concatenated anchor key and value vectors.  To edit any frame $I^t \in \mI$, we modify the self-attention module to the cross-frame attention with respect to the key and value vectors of anchor frames as follows:
\begin{align*}
    \softmax \left( \frac{\mQ [\mK, \mK_{anc}]^T}{\sqrt{d}} \right) [\mV, \mV_{anc}],
\end{align*}
where $\mQ, \mK$ and $\mV$ are the self-attention vectors of the input frame $I^t$. The idea is that the attention score generated by the softmax facilitates cross-frame tracking by estimating the temporal correspondence between the input frame and anchor frames. The global value vectors then be propagated to input frame by multiplying the attention score with $\mV_{anc}$. By substituting the self-attention module with an anchor-based cross-frame attention mechanism, we found that the model could generate highly consistent video edits. In the default setting, we choose anchor frames uniformly across the video, and we did not notice consistent performance improvement or degradation when adopting different anchor-frame selection strategies.

\paragraph{Fast Generation via Parallelization}
Note that editing frame $I_t$ does not require other frames as input except the cached features $\mK_{anc}$ and $\mV_{anc}$ from anchor frames. Therefore, we can edit arbitrary long videos by splitting them into segments and leverage multi-GPUs to parallize the generation, while the computation remains numerically identical. As a result, our method achieves significant speedup compared to previous works. Moreover, it delivers superior quality outputs without succumbing to memory-related constraints. This efficiency underscores our approach's enhanced scalability and practicality, setting a new benchmark for performance in the realm of video editing.

\subsection{Equivariant Finetuning}
\label{sec:equi-ft}

While anchor-based attention greatly improves the quality, we still occasionally observed temporal inconsistency. In particular, we found that for generated contents that do not have semantic correspondence with the input, small changes in input frames can cause significant variances in the output frames.

\begin{figure*}[t!]
\begin{center}   
\includegraphics[width=\linewidth]{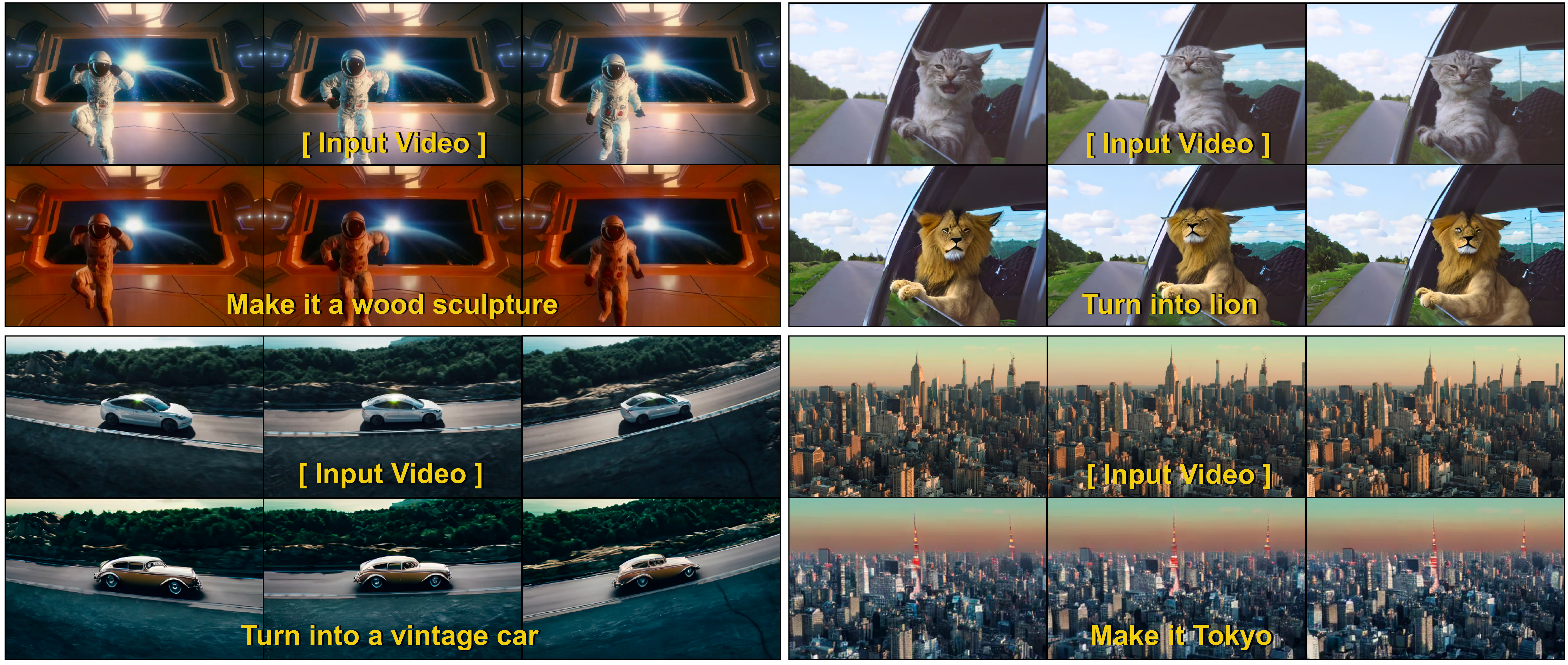}
\end{center}
\vspace{-6mm}
\caption{\textbf{Diverse Video Editing via \ourmodel.} \ourmodell enables a wide range of video edits with different types of subjects. 
} 
\vspace{-3mm}
\label{fig:exp_1}
\end{figure*}

To improve the consistency, we leverage the following intuition to design a data augmentation strategy. In particular, if an input frame $I^t$ differs from $I^{t-1}$ only in the camera position, then the output frame $\hat{I}^t$ and $\hat{I}^{t-1}$ should only be different in the camera position as well. This inspires us to come up with a data augmented strategy that can be applied to any image editing dataset to imporve the temporal consistency. Given a pair of images, the original and the edited, denoted as $(I, I')$, we randomly sample an affine transformation $g: \gI \rightarrow \gI$ and apply them to both images to obtain $(g(I), g(I'))$. We implement this using torchvision's random affine transformation \cite{torchvision}, setting random rotations degrees to $<5^\circ$, random translation to $[-0.05, 0.05]$, random scaling factor to $[0.95, 1.05]$, and random shear degrees to $[-5^\circ, 5^\circ]$ on both axis. We also apply random resized crop, scaling the original image to 288pix, and randomly crop a square image with 256 pix. We then fine-tuned the base image-to-image model to generate the transformed $g(I')$ given the transformed $g(I)$. The proposed fine-tuning process makes the model equivariant to affine transformations, leading us to denote our approach as \emph{equivariant finetuning}. Empirically, we observe a notable enhancements in temporal consistency after finetuning (Section \ref{sec:ablation}).

%% file: sec/6_experiments.tex
\section{Results}
We implement \ourmodell based on an instruction-based image editing model, similar to \cite{brooks2023instructpix2pix}, and replace the model's self-attention with cross-frame attention. We set the number of anchor frames to 3. Anchor frames are uniformly selected with equal intervals among all frames. The model can accept input with different aspect ratios, and we rescale the input resolution with the longer size to be 512, and keep the aspect ratio unchanged. We edit all frames of the input video, without temporal downsampling. We distribute the computation to 8 A100 GPUs. We use the Euler Ancestral sampler with 10 diffusion steps. 

For equivariant finetuning, we use the same dataset that was used to train the image editing model, and apply the data augmentation discussed in Section \ref{sec:equi-ft}. We load the image editing model's pretrained checkpoint, and resumed training for 50,000 steps with a batch size of 128, costing 30 hours on 8 A100 GPUs with 80GB memories.

\subsection{Qualitative Evaluation}
\begin{figure}[!t]
\begin{center}   \includegraphics[width=0.95\linewidth]{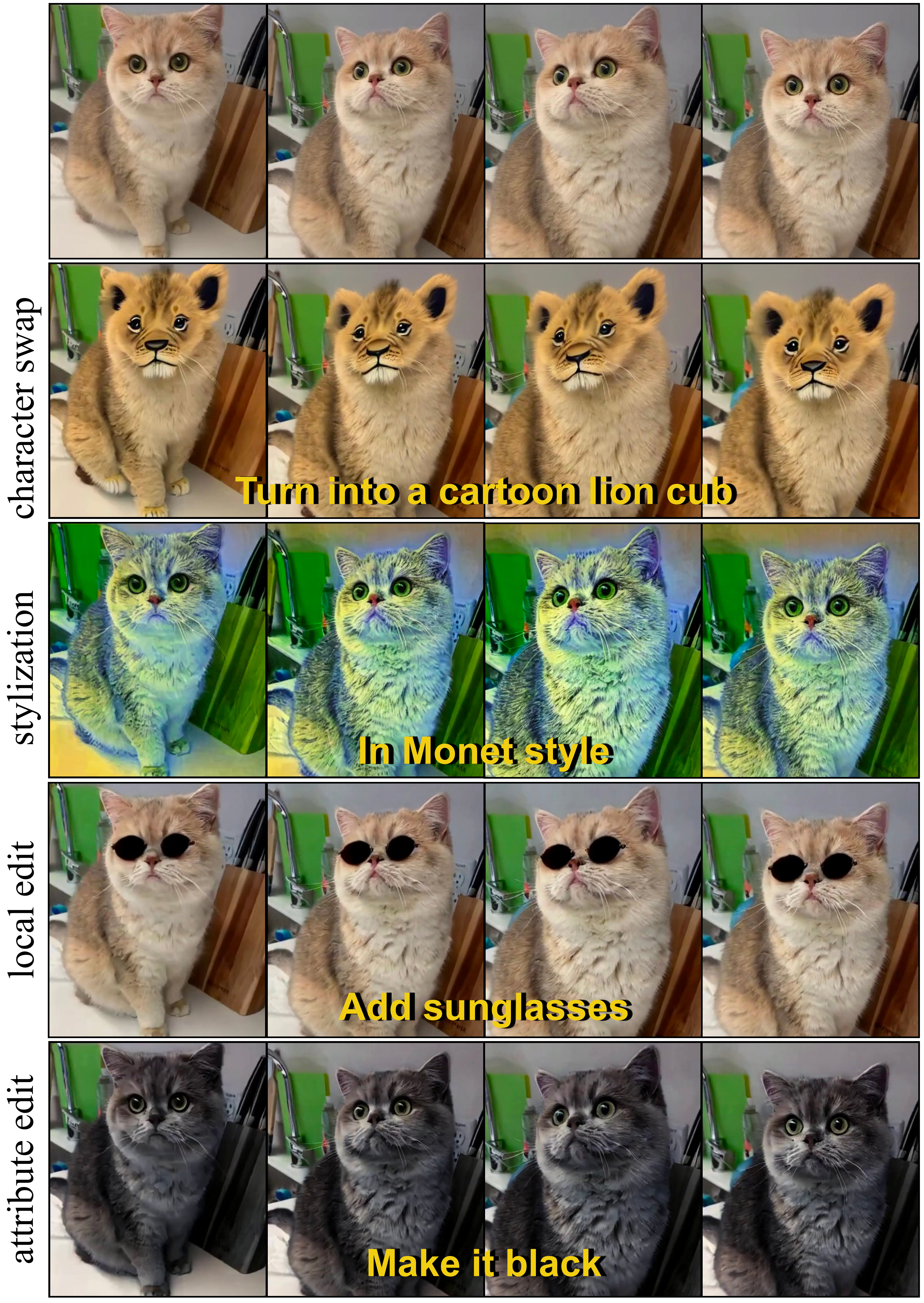}
\end{center}
\vspace{-6.5mm}
   \caption{\textbf{Different type of editing.} \ourmodell is able to handle a diverse set of instructions and perform appropriate editing. } 
\label{fig_exp_2}
\vspace{-2.5mm}
\end{figure}

\begin{figure*}[t!]
\begin{center}   
\includegraphics[width=0.962\linewidth]{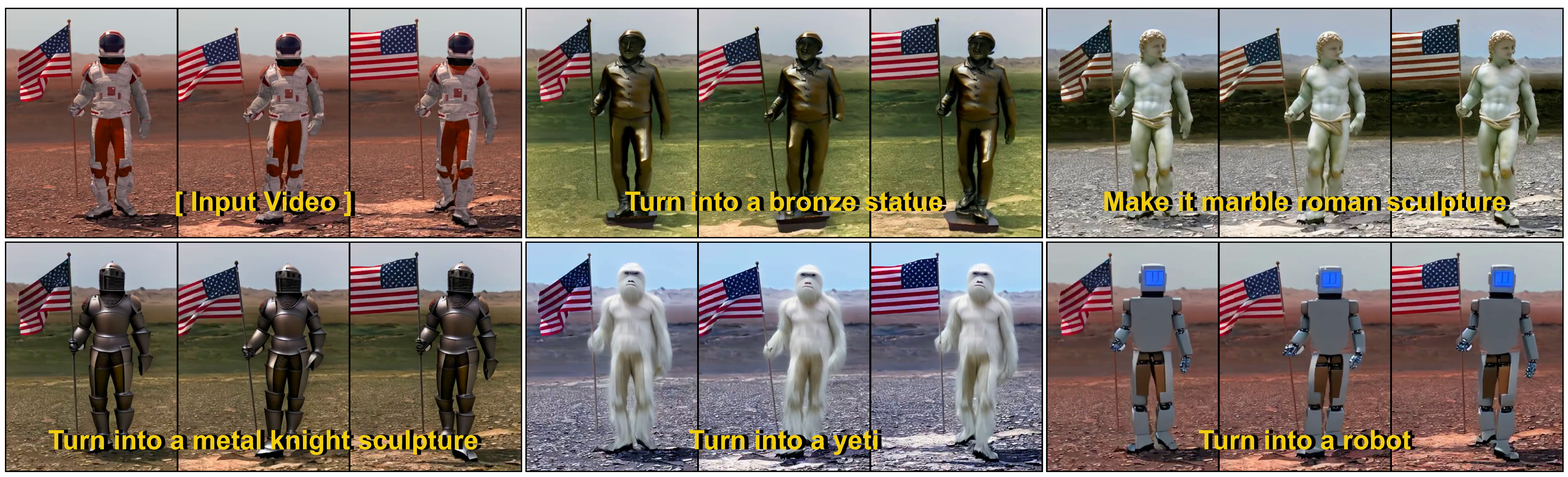}
\end{center}
\vspace{-5mm}
\caption{\textbf{Diverse Character Swap:} \ourmodell possesses the capability to interchange the individual with a diverse array of characters. 
} 
\vspace{-1mm}
\label{fig:exp_charswap}
\end{figure*}

\vspace{-2mm}
We first show qualitative results of \ourmodel. Since most of the PDF readers do not render videos properly, we only show a small number of frames for each video. We strongly recommend readers to checkout our supplementary materials to watch the complete videos. In Figure \ref{fig:exp_1}, we show that our model is capable conducting edits on different subjects. In Figure \ref{fig_exp_2}, we show that our model is able to conduct different types of editing, including stylization, character swap, local editing, and attribute editing, following textual instructions. 
In Figure \ref{fig:exp_charswap}, we show that our model can transform the source character into different target characters based on instructions. Note that our model can adapt to different input aspect ratios without need for re-training. Our input videos contain large motions, occlusions, and other complex dynamics. Despite those challenges, videos generated by our model are temporally consistent and visually appealing. We also show our model's capabilities to generate long videos in the supplementary material.

\subsection{Quantitative Evaluation}
\label{sec:quant}
\vspace{-2mm}
Quantitatively evaluating video generative models is challenging. First, the generation task is intrinsically high-variance -- there are countless ways to edit an video given the instruction.
Second, previous works have adopted metrics such as CLIP scores \cite{esser2023structure,geyer2023tokenflow} to evaluate the generation quality. However, these metrics are not necessarily aligned with human perception \cite{liu2023evalcrafter}. Lastly, human evaluation is still the golden standard to judge the quality. Yet, due to the cost of human evaluation, previous works have only conducted small scale human evaluations ($<100$ samples). 

In this paper, we conduct a large-scale user study on an evaluation set consists of 1000 video-instruction samples. The evaluation set is divided into two parts: first, to test a model's robustness across different videos, we construct the evaluation set of 50 videos $\times$ 10 instructions. And to test a model's robustness across different instructions, we construct a dual evaluation set of 10 videos $\times$ 50 instructions. The videos are accessible from ShutterStock \cite{ShutterstockVideo}. To our best knowledge, this is the largest evaluation in the video-to-video generation literature so far.

\begin{figure}[t]
\begin{center}   \includegraphics[width=0.95\linewidth]{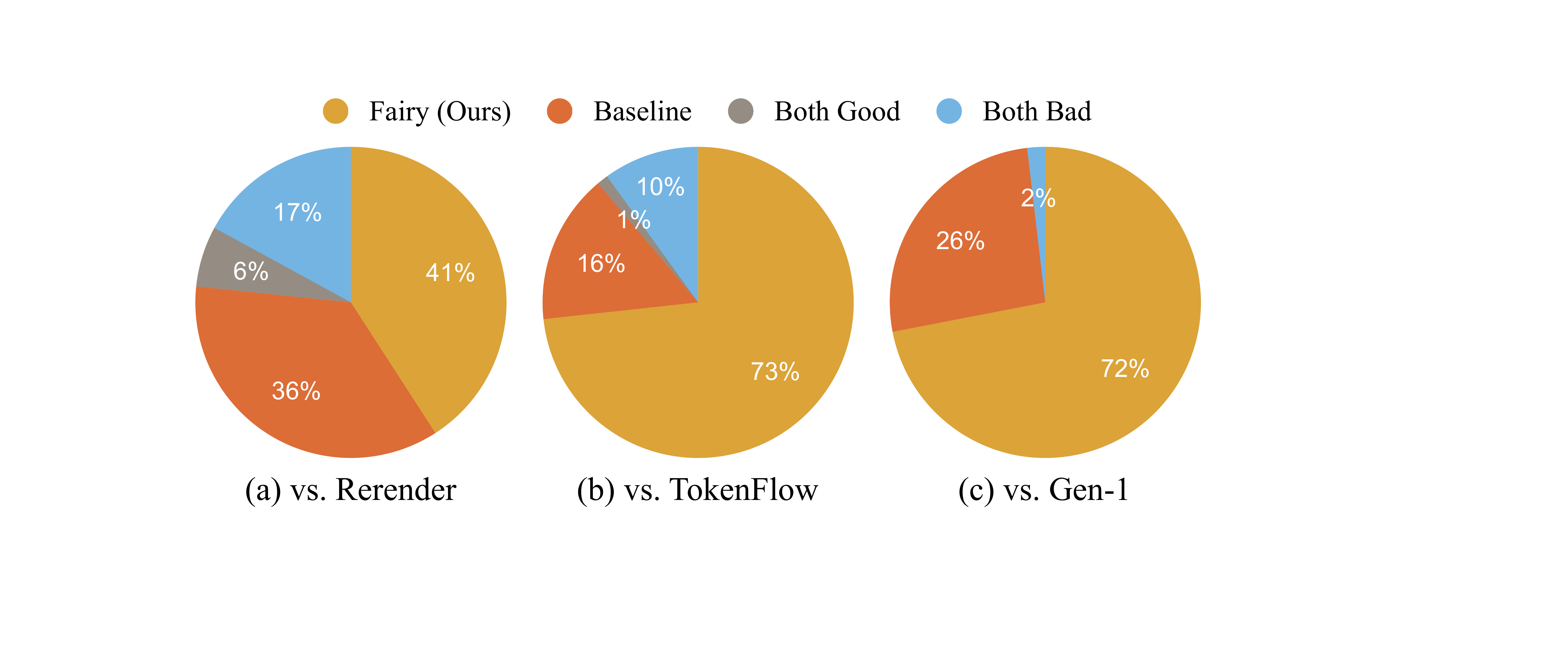}
\end{center}
\vspace{-4.5mm}
   \caption{\textbf{A/B Comparison with Baselines.} \ourmodell significantly surpassed baseline models, demonstrating its effectivity. } 
\label{fig_exp_ab}
\vspace{-1.5mm}
\end{figure}

We conduct a A/B comparison to compare our method with three previous works, Rerender \cite{yang2023rerender} (tracking and propagation), Tokenflow \cite{torchvision} (image model adaptation), and Gen-1 \cite{esser2023structure} (conditional video model), which are the strongest representative of the three paradigms for video-to-video generation today. Results from baselines are collected from \cite{comparison}. Prompts for baselines are descriptive, e.g., "a dog running on grass, in Van Gogh style". We re-write the prompt for our method as an edit instruction, e.g., "in Van Gogh Style". Since Gen-1 is not open-sourced, the evaluation is done on a smaller evaluation set of 100 videos. In each evaluation tuple, we show the input video, the editing instruction or prompt, and the output videos generated by \ourmodell and a baseline. We ask human evaluators to choose the better video in terms of their single frame quality, temporal consistency, prompt faithfulness, input faithfulness, and overall quality. Each comparison is rated by 3 different annotators
and the decision is determined by the majority vote. We report the overall quality comparison in Figure \ref{fig_exp_ab}, which demonstrates that videos produced by \ourmodell are more preferable, with a win rate of 41\% vs 36\% against Rerender, 73\% vs. 16\% against TokenFlow, and 72\% vs 26\% against Gen-1. More details in the supplementary material.

\begin{table}[t!]
\small
\centering
\renewcommand{\tabcolsep}{1.6pt}
\begin{tabular}{@{}l|c|c|c}
& Latency (sec) $\downarrow$ & Frame-Acc $ \uparrow$ & Tem-Con $\uparrow$ \\ 
\hline
\hline
TokenFlow \;\;& 744 & 0.537 & 0.973  \\
Rerender & 608 & 0.775 & 0.972  \\
\hline
\textbf{Ours} &  \textbf{13.8}  &  \textbf{0.819} & \textbf{0.974}    \\
\hline
\end{tabular}
\vspace{-2.5mm}
\caption{We assess our method's temporal consistency and fidelity to the target text prompt using CLIP similarity metrics.}
\vspace{-2mm}
\label{tab:comparison}
\end{table}

\begin{figure}[t!]
\begin{center}   \includegraphics[width=\linewidth]{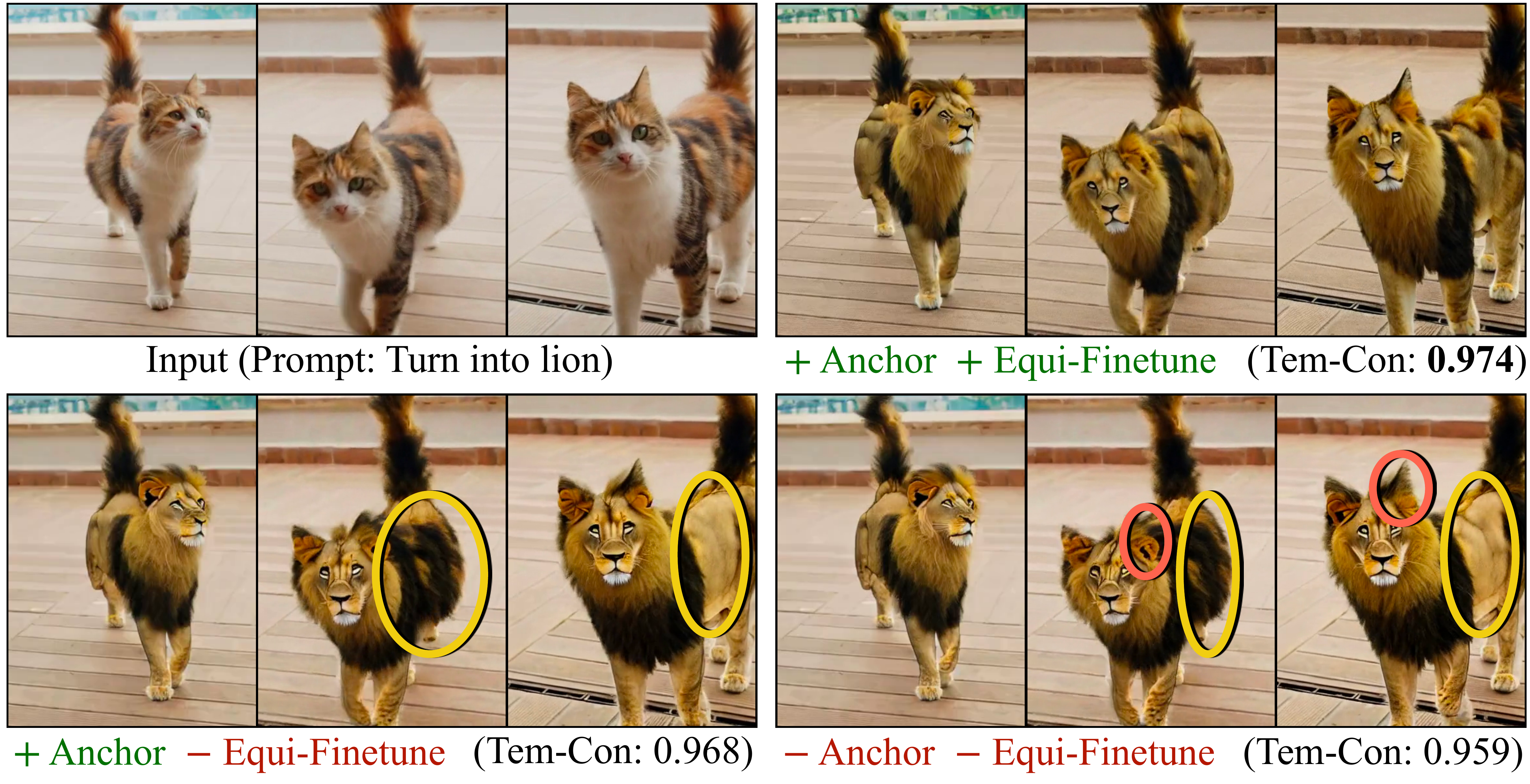}
\end{center}
\vspace{-5mm}
   \caption{\textbf{Ablation Study.} Without equivariant fine-tuning and anchor-based attention, we observed inconsistencies, particularly in the changing patterns of body and earscostumes over time. This inconsistency is further exacerbated upon the removal of anchor-based attention, leading to lower temporal consistency score.
   } 
\label{fig_exp_ablation}
\vspace{-2mm}
\end{figure}

\begin{figure*}[t!]
\begin{center}   
\includegraphics[width=0.95\linewidth]{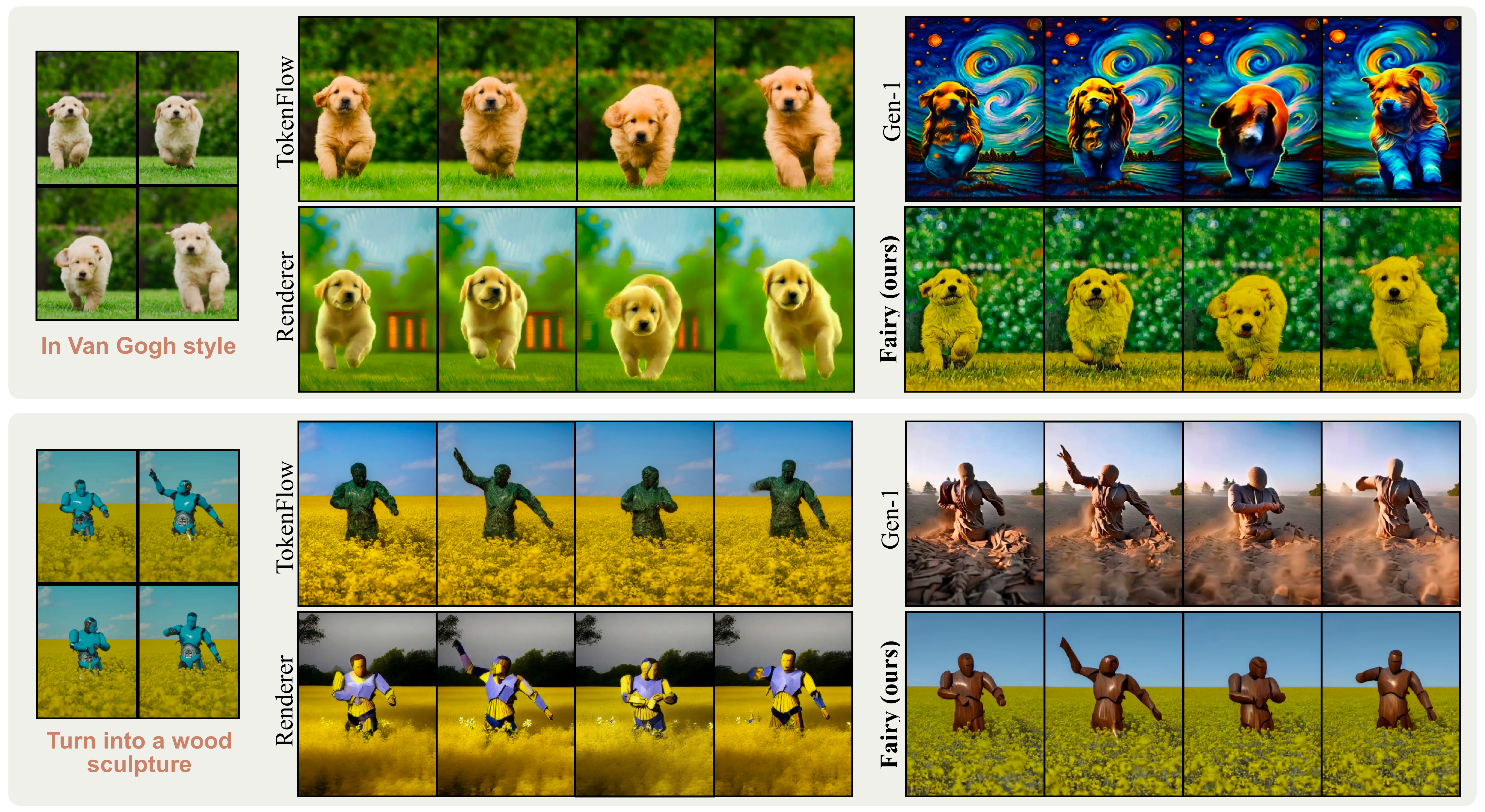}
\end{center}
\vspace{-6mm}
\caption{\textbf{Comparison with Baselines.} \ourmodell consistently outperform baselines in terms of consistency and instruction-faithfulness. 
} 
\vspace{-3.5mm}
\label{fig:exp_compare}
\end{figure*}

Figure \ref{fig:exp_compare} shows the visual comparison with the baselines. We observe that both Tokenflow \citep{geyer2023tokenflow} and Rerender \citep{yang2023rerender} do not adhere closely to the provided instructions, resulting in evident inconsistencies. Outputs from Gen-1 often over-modify the entire scene and do not retain the original content effectively. In contrast, \ourmodell meticulously follows the instruction, delivering high-quality, temporal consistent, and authentic generations.

Lastly, we compute metrics adopted by previous works, Tem-Con and Frame-Acc \citep{qi2023fatezero, yang2023rerender}. Tem-Con assesses temporal consistency by calculating the cosine similarity of CLIP feature across successive frame pairs, and Frame-Acc measures the percentage of frames where the edited image exhibits greater CLIP similarity to the target prompt than to the source prompt. The results in Table \ref{tab:comparison} demonstrates that \ourmodell achieves the best temporal consistency and frame-wise editing accuracy against the baselines.

\paragraph{Speed Comparison} In Table \ref{tab:comparison}, we also compare the latency of different models. In particular, we calculate the inference time of editing a 4-seconds, 30 FPS, 512p$\times$384p video on a server with 8 A100 GPUs. The key-frame interval of Rerender is set to 4 instead of the default 10, since the test videos contain faster motion. This leads to improved quality for Rerender. All other parameters were default. Due to its architecture simplicity, \ourmodell is already significantly faster than baselines using 1 GPU. Using a single GPU, \ourmodell completes inference in just 78 seconds, achieving 9.5$\times$ faster than TokenFlow and 7.5$\times$ faster than Rerender. When utilizing all 8 GPUs on the node, \ourmodell is 53$\times$ faster than TokenFlow and 44$\times$ faster than Rerender.

\subsection{Ablation Studies}
\label{sec:ablation}
We conduct an ablation study to verify the effectiveness of our model's component. We gradually remove equivariant fine-tuning and anchor-based attention, ultimately leading to the adoption of a standard frame-by-frame editing approach. The results are shown in Figure \ref{fig_exp_ablation}. The model becomes sensitive to the camera motion without equivariant finetuning, rendering inconsistency in the details. The subsequent removal of anchor-based attention, transitioning to a frame-based model, introduces further inconsistencies in the generated video. We compute the Tem-Con metric based on 150 videos and report in Figure \ref{fig_exp_ablation}. It confirms our observation that the proposed methodology effectively improves the temporal consistency, lifting the Tem-Con from 0.959 (baseline) to 0.968 (w/ anchors) to 0.974 (w/ anchor and equivariant finetuning).

\subsection{Limitations}
\label{sec_limitations}
The efficacy of \ourmodell is intrinsically tied to the underlying image-editing model. This means that any inherent constraints of this underlying model, e.g., face and text distortion, etc., will naturally manifest in the video editing capacities of \ourmodell. In our observations, a notable side effect of equivariant finetuning is the diminished ability to accurately render dynamic visual effects, such as lightning or flames. The process seems to overly focus on maintaining temporal consistency, which often results in the depiction of lightning as static or stagnate, rather than dynamic and fluid. See the supplementary material for visualization.

%% file: sec/7_conclusion.tex
\section{Conclusion}

\ourmodell offers a transformative approach to video editing, building on the strengths of image-editing diffusion models. By leveraging anchor-based cross-frame attention and equivariant finetuning, \ourmodell guarantees temporal consistency and superior video synthesis. Moreover, it tackles the memory and processing speed constraints observed in preceding models. With the capability to produce high-resolution videos at a blazing speed, \ourmodell firmly establishes its superiority in terms of quality and efficiency, as further corroborated by our extensive user study.

%% file: sec/appendix.tex
\newpage
~
\newpage
\section{Webpage Demo}
The videos in the main paper and appendix can be viewed with our demo webpage by opening the \url{webpage/index.html} in the supplementary material using a web browser.

\section{Additional Results from Human Evaluation}
\begin{figure}[h]
\begin{center}   \includegraphics[width=1.0\linewidth]{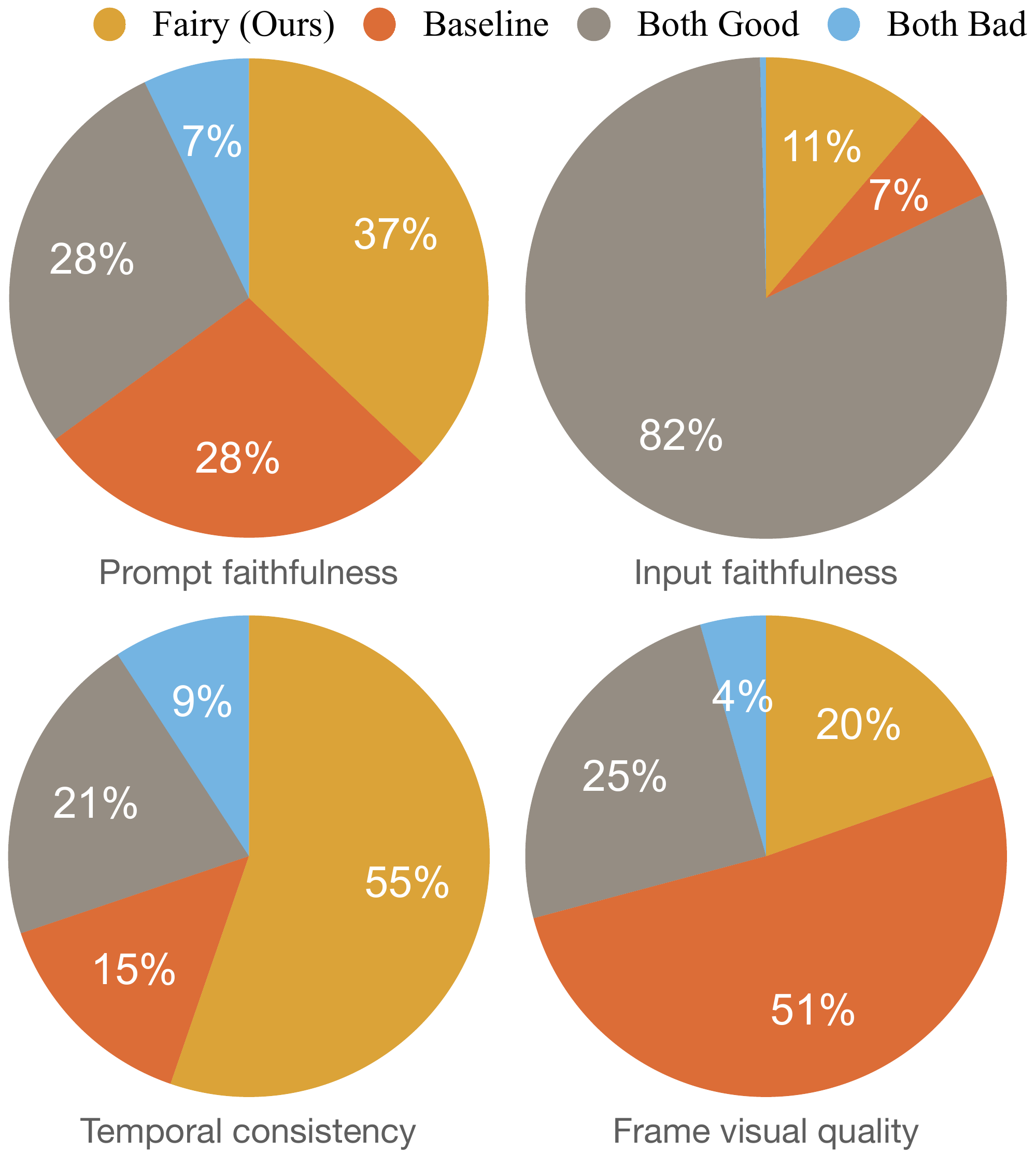}
\end{center}
   \caption{\textbf{Comparison with Rerender.} } 
\label{fig_vs_rerender}
\end{figure}

\begin{figure}[h]
\begin{center}   \includegraphics[width=1.0\linewidth]{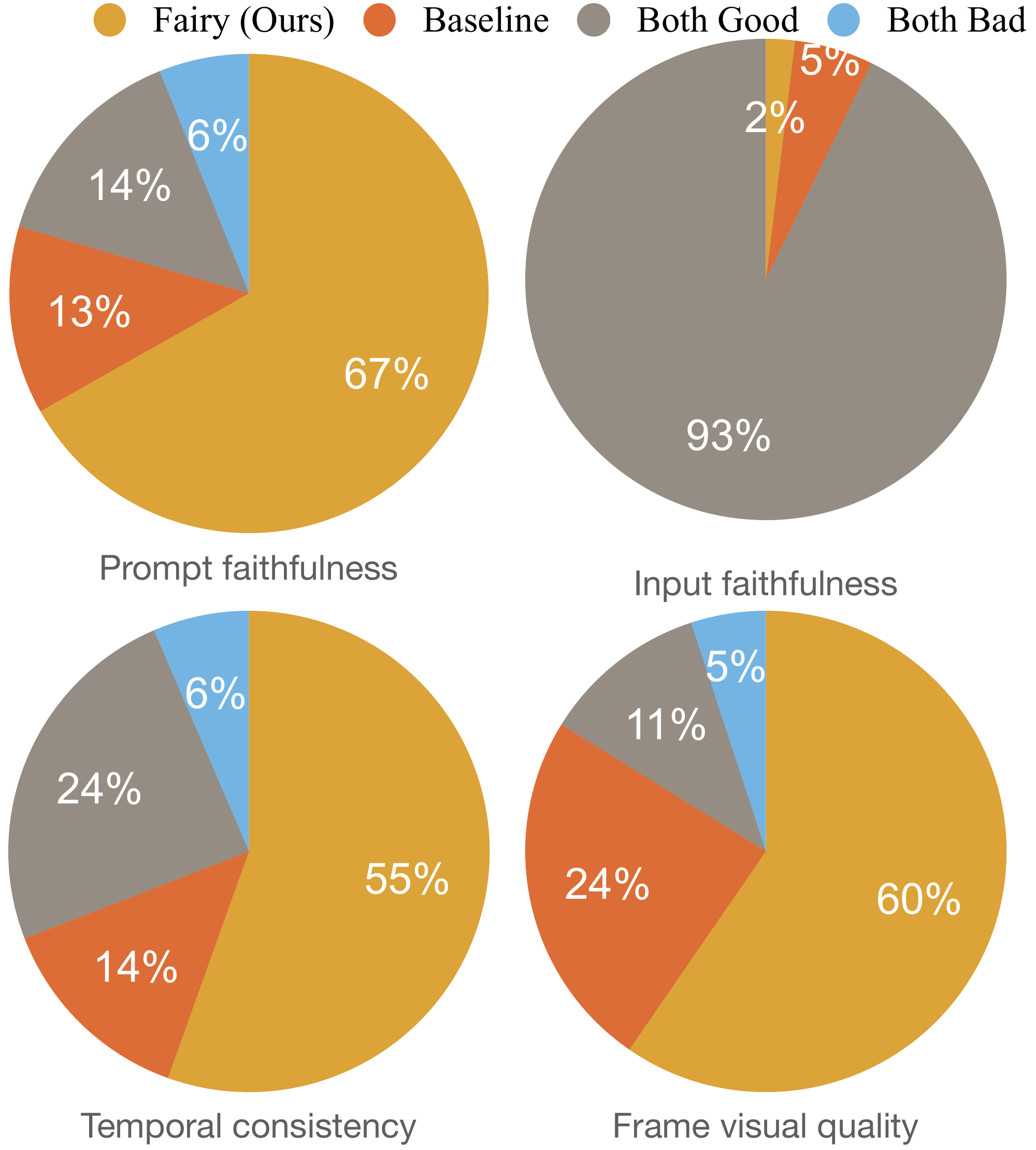}
\end{center}
   \caption{\textbf{Comparison with TokenFlow.}} 
\label{fig_vs_tokenflow}
\end{figure}

\begin{figure}[h]
\begin{center}   \includegraphics[width=1.0\linewidth]{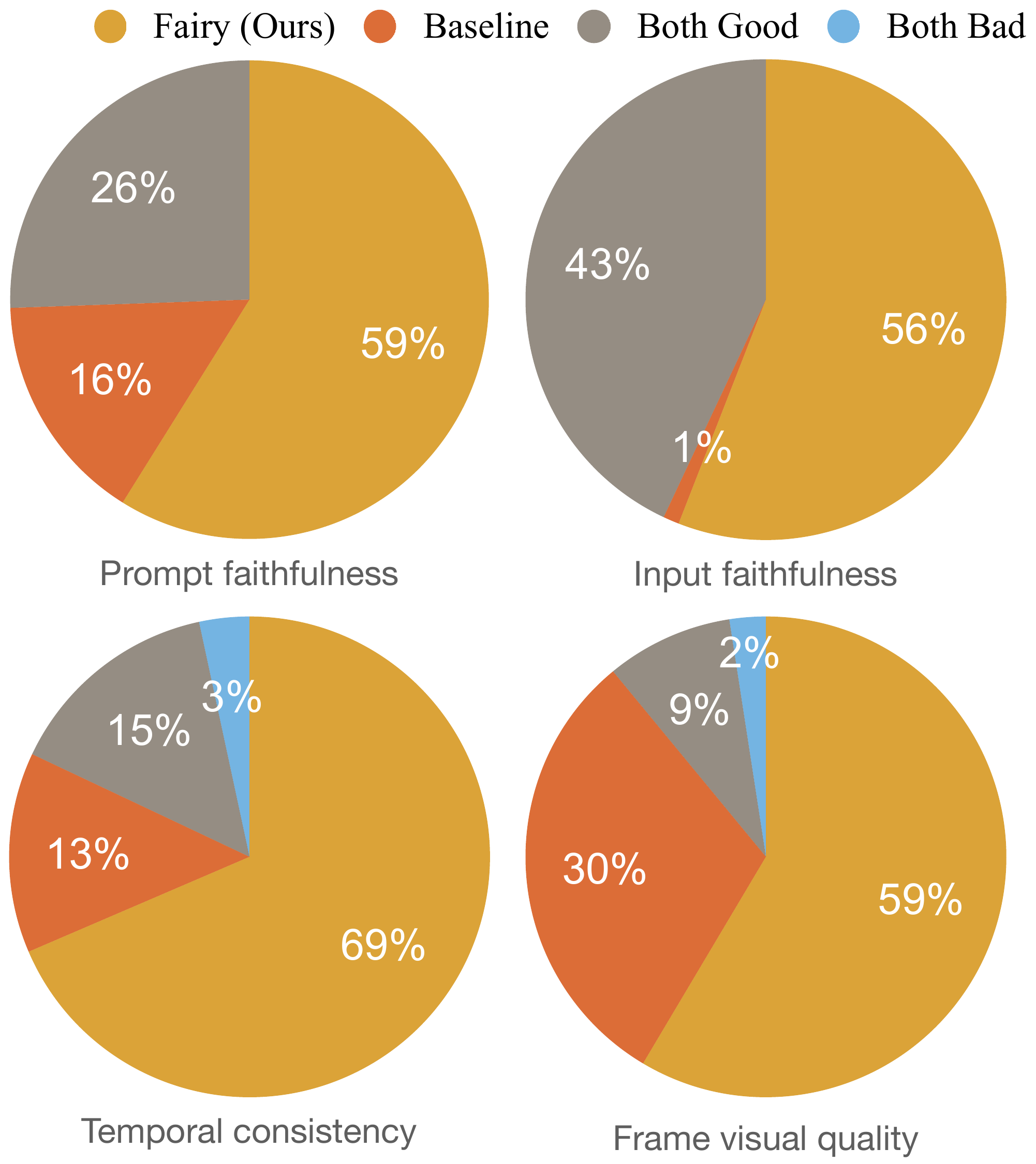}
\end{center}
   \caption{\textbf{Comparison with Gen-1.} } 
\label{fig_vs_runway}
\end{figure}

As mentioned in Section \ref{sec:quant}, we conduct A/B comparison between \ourmodell with baselines. We ask annotators to compare our generated video with a baseline method's result, and decide which one is better.Each video pair is evaluated by three independent annotators, and the majority vote is considered as the final rating. We ask raters to evaluate by four attributes: frame quality -- visual quality of single frame; temporal consistency -- whether the frames are coherent or flickering; prompt faithfulness -- whether the output followed the editing instruction or target prompt; input faithfulness -- whether the output video followed the contents of the original video. We reported the overall rating in Section \ref{sec:quant}, Figure \ref{fig:exp_compare}. Here, we report a more detailed comparison along each attributes in Figure \ref{fig_vs_rerender}, \ref{fig_vs_tokenflow}, \ref{fig_vs_runway}. Compared with Rerender (Figure \ref{fig_vs_rerender}), \ourmodell loses in terms of single frame visual quality. This is mainly due to the limitation of the foundational image editing model, while Rerender utilizes LoRA to enhance frame quality. Yet, \ourmodell significantly outperforms Rerender in terms of temporal consistency, achieves better prompt faithfulness, and performs similarly in terms of input faithfulness. Compared with TokenFlow, \ourmodell outperforms significantly in terms of frame quality, temporal consistency, and prompt faithfulness. They performs similarly in terms of input faithfulness. Compared with Gen-1, \ourmodell significantly outperforms in all attributes.

\begin{figure}[h]
\begin{center}   \includegraphics[width=1.0\linewidth]{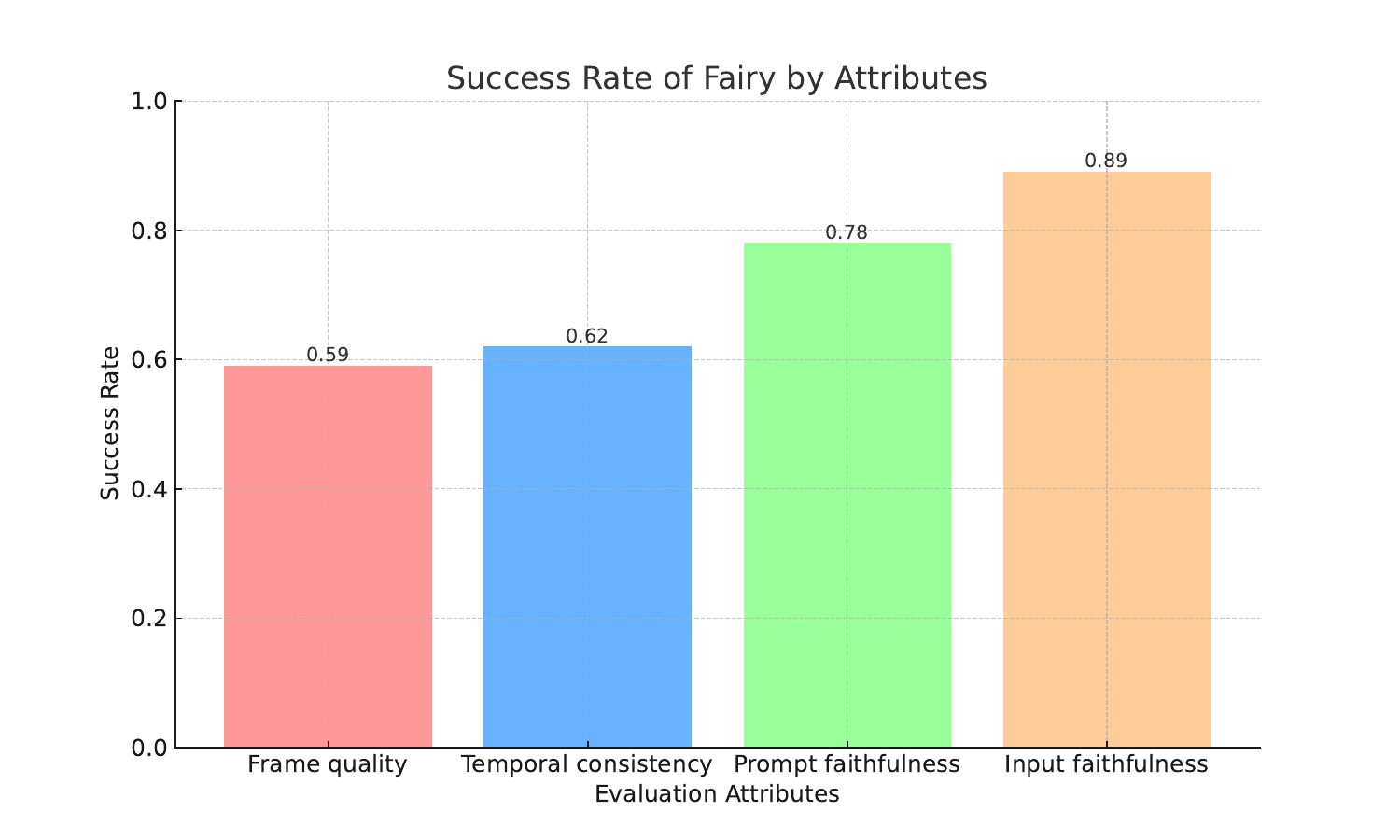}
\end{center}
\vspace{-6.5mm}
   \caption{\textbf{Standalone success rate by attributes.} We report \ourmodel's success rate in terms of frame quality, temporal consistency, prompt faithfulness, and input faithfulness.} 
\label{fig_standalone_attrib}
\vspace{-2.5mm}
\end{figure}

In addition to the A/B comparison, in which we ask human annotators to compare our method with a baseline, we also conduct a standalone evaluation to examine output video's quality. Each time we show an annotator the original video, an editing instruction, and the result video. We then ask the annotator to rate the output as good or bad by the same four attributes.  We ask 3 annotators to rate each video, and the decision is determined by their majority vote. We report the success rate by each attributes in Figure \ref{fig_standalone_attrib}.

\section{More Results}

\subsection{Character Swap}

\begin{figure*}[t!]
\begin{center}   
\includegraphics[width=0.962\linewidth]{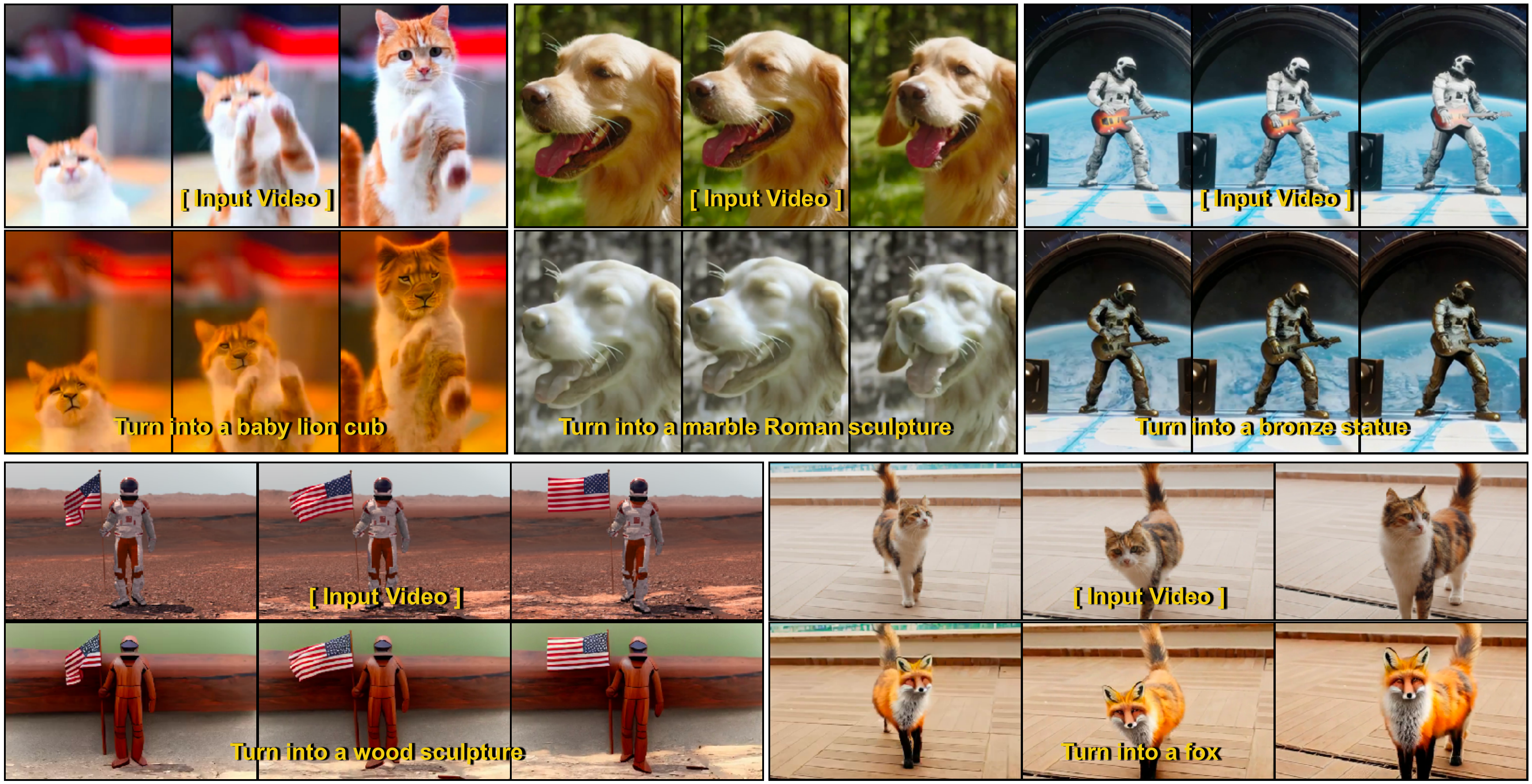}
\end{center}
\vspace{-5mm}
\caption{\textbf{Additional Results on Character Swap:} \ourmodell is able to interchange the characters for videos with arbitrary ratios.
} 
\vspace{-1mm}
\label{fig_appendix_charswap}
\end{figure*}

In Figure \ref{fig_appendix_charswap}, we demonstrate more results of character swap, where \ourmodell is able to interchange individuals with various characters. Note that our model can adapt to different input aspect ratios without need for re-training.

\subsection{Stylization}

\begin{figure*}[t!]
\begin{center}   
\includegraphics[width=0.962\linewidth]{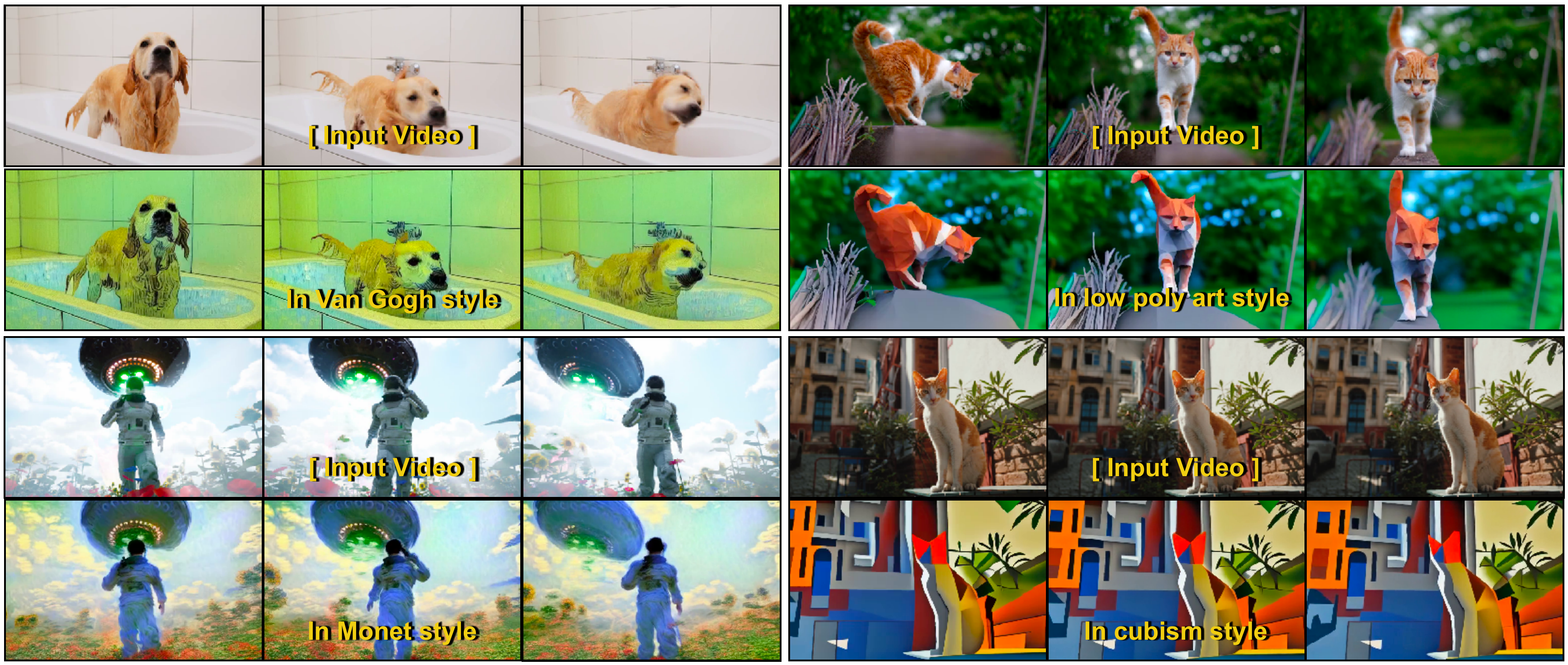}
\end{center}
\vspace{-5mm}
\caption{\textbf{Additional Results on Stylization:} \ourmodell enables a wide range of style editing.
} 
\vspace{-1mm}
\label{fig_appendix_stylization}
\end{figure*}

Figure \ref{fig_appendix_stylization} demonstrates more stylization results of \ourmodel. In particular, our model is able to recognize various styles, while perform high quality and temporal consistent edit based on the stylization instructions.

\subsection{Arbitrary Long Videos}
\begin{figure*}[t!]
\begin{center}   
\includegraphics[width=0.962\linewidth]{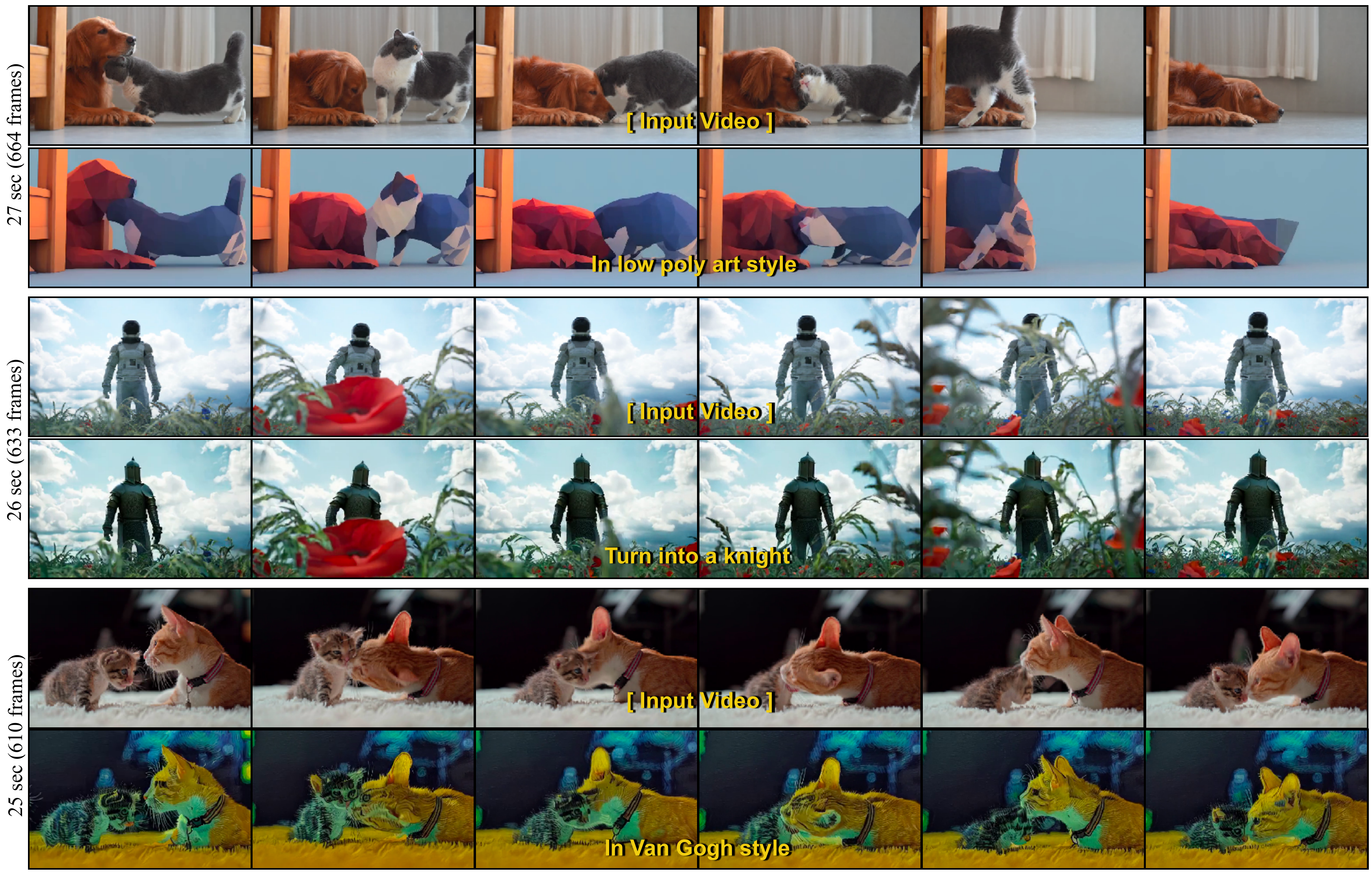}
\end{center}
\vspace{-5mm}
\caption{\textbf{Any-length Video Editing.} \ourmodell is able to scale to arbitrary long video without memory issue.
} 
\vspace{-1mm}
\label{fig_appendix_longvid}
\end{figure*}

\ourmodell is able to scale to arbitrary long video without memory issue due to the proposed anchor-based attention. In Figure \ref{fig_appendix_longvid}, we show that our model is able to generate a 27 second long video with high quality, while the latency is less than 71.89 seconds via 6 A100 GPUs. In particular, the \ourmodell manage to retain decent temporal consistency even number of frames (664 frames) is way more than the number of anchor frames (3 frames).

\subsection{Ablation Study}

\begin{figure*}[t!]
\begin{center}   
\includegraphics[width=0.962\linewidth]{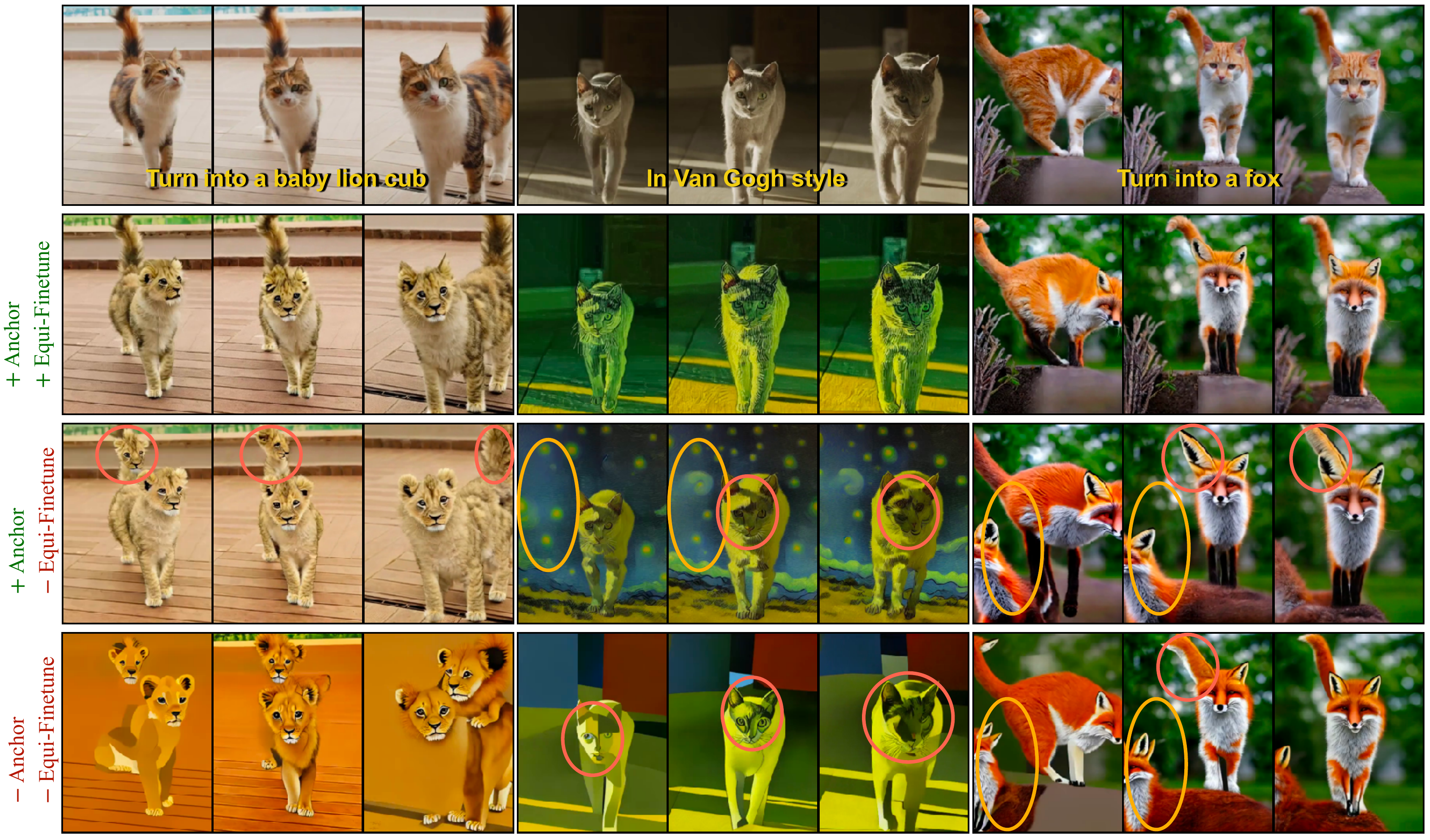}
\end{center}
\vspace{-5mm}
\caption{\textbf{Additional Results on Ablation Study.} We demonstrate that both equivariant finetuning and anchor-based attention are crucial to \ourmodell.
} 
\vspace{-1mm}
\label{fig_appendix_ablation}
\end{figure*}

Figure \ref{fig_appendix_ablation} shows more ablation results by removing equivariant finetuning and anchor-based attention. We can see that without equivariant finetuning, the model is sensitive to local motion and movement of the subject and therefore degenerate the frame quality and temporal consistency. For instance, in the first video, the tail of the cat becomes the head of the lion in some of the frames, and the face of the cat in second video vary significantly between frames. Without anchor-based attention, the edit of each frame is completely independent, rendering in significant worse temporal consistency.

\begin{figure*}[t!]
\begin{center}   
\includegraphics[width=0.962\linewidth]{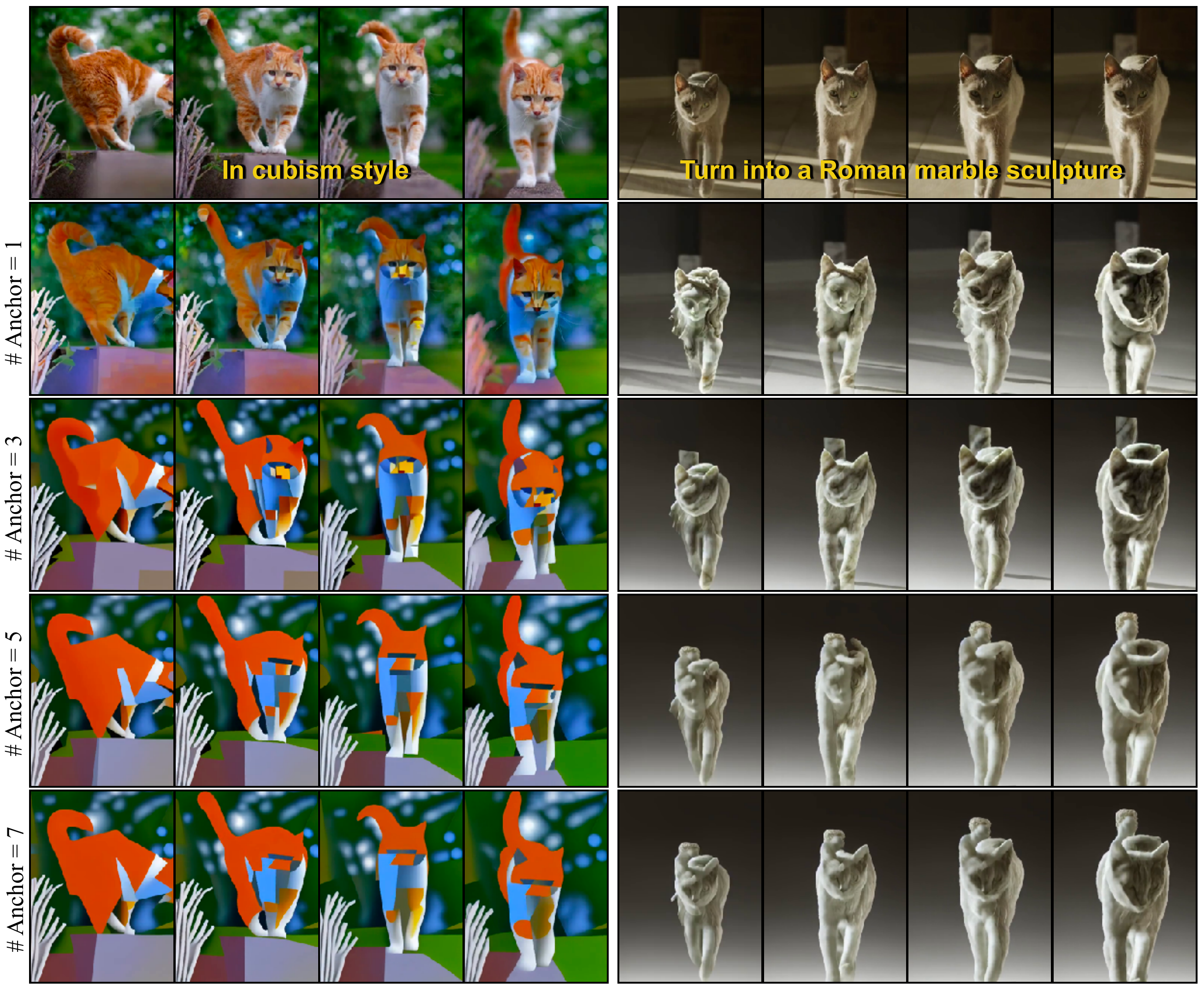}
\end{center}
\vspace{-5mm}
\caption{\textbf{Ablation study on number of anchor frames.} We found that setting number of anchor frames to 3 yields the best results.
} 
\vspace{-1mm}
\label{fig_appendix_anchor}
\end{figure*}

Figure \ref{fig_appendix_anchor} demonstrates results generated with different number of anchor frames. When number of anchor frames equals to 1, the global features model can leverage are too restricted, which lead to suboptimal edits. In contrast, we observe that when the number of anchor frames is greater than 7, the quality also gradually degrades, losing some visual details. 

\begin{figure*}[t!]
\begin{center}   
\includegraphics[width=0.962\linewidth]{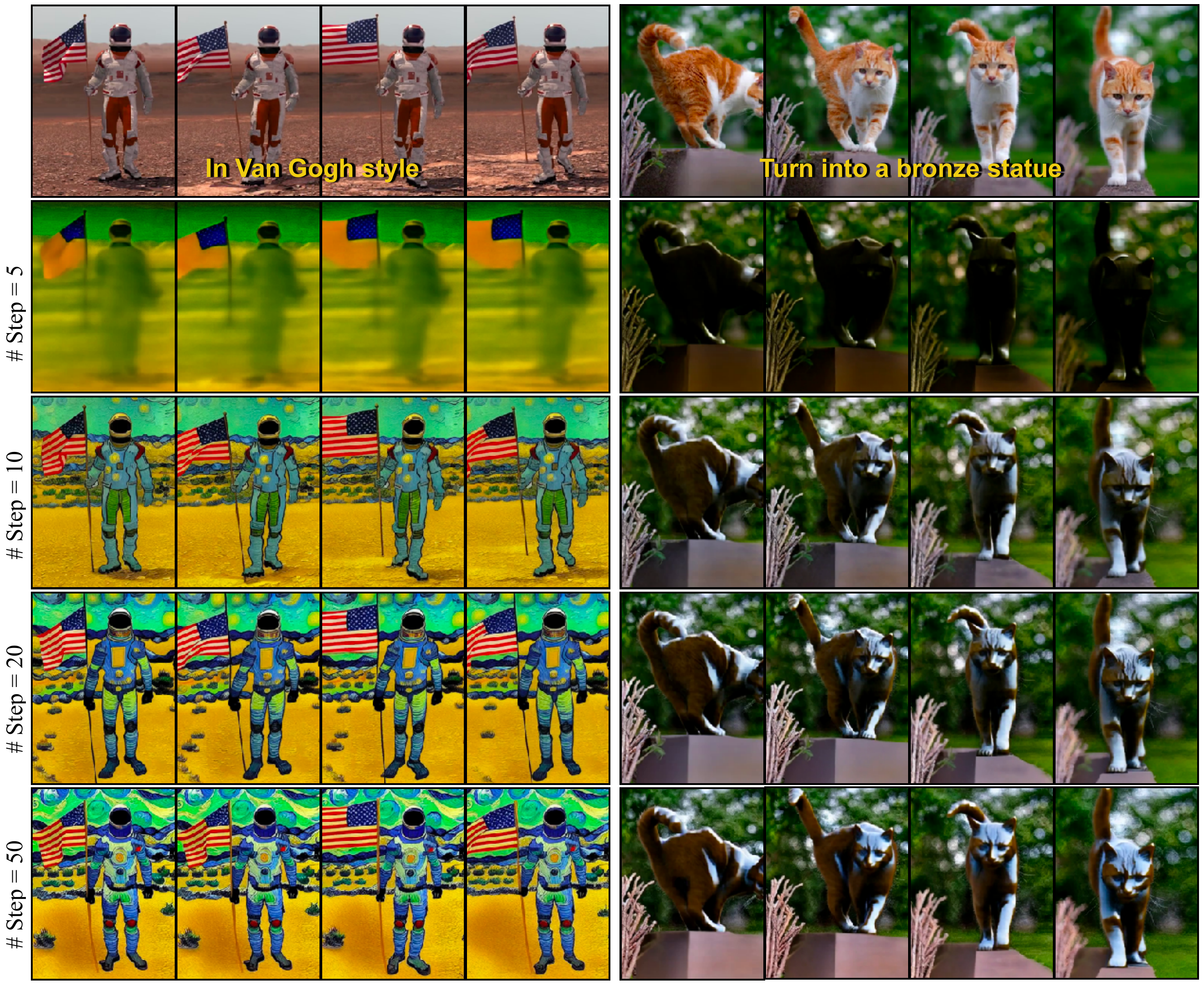}
\end{center}
\vspace{-5mm}
\caption{\textbf{Ablation study on number of diffusion steps.} We found that diffusion steps above 5 generally yield good results.
} 
\vspace{-1mm}
\label{fig_appendix_steps}
\end{figure*}

In Figure \ref{fig_appendix_steps}, we perform ablation study on the number of diffusion steps during generation. The model perform reasonably well when the number of diffusion step is above 10. We therefore set the diffusion step to 10 for all of our experiments to optimize the latency.

\subsection{Limitations}

\begin{figure*}[t!]
\begin{center}   
\includegraphics[width=0.962\linewidth]{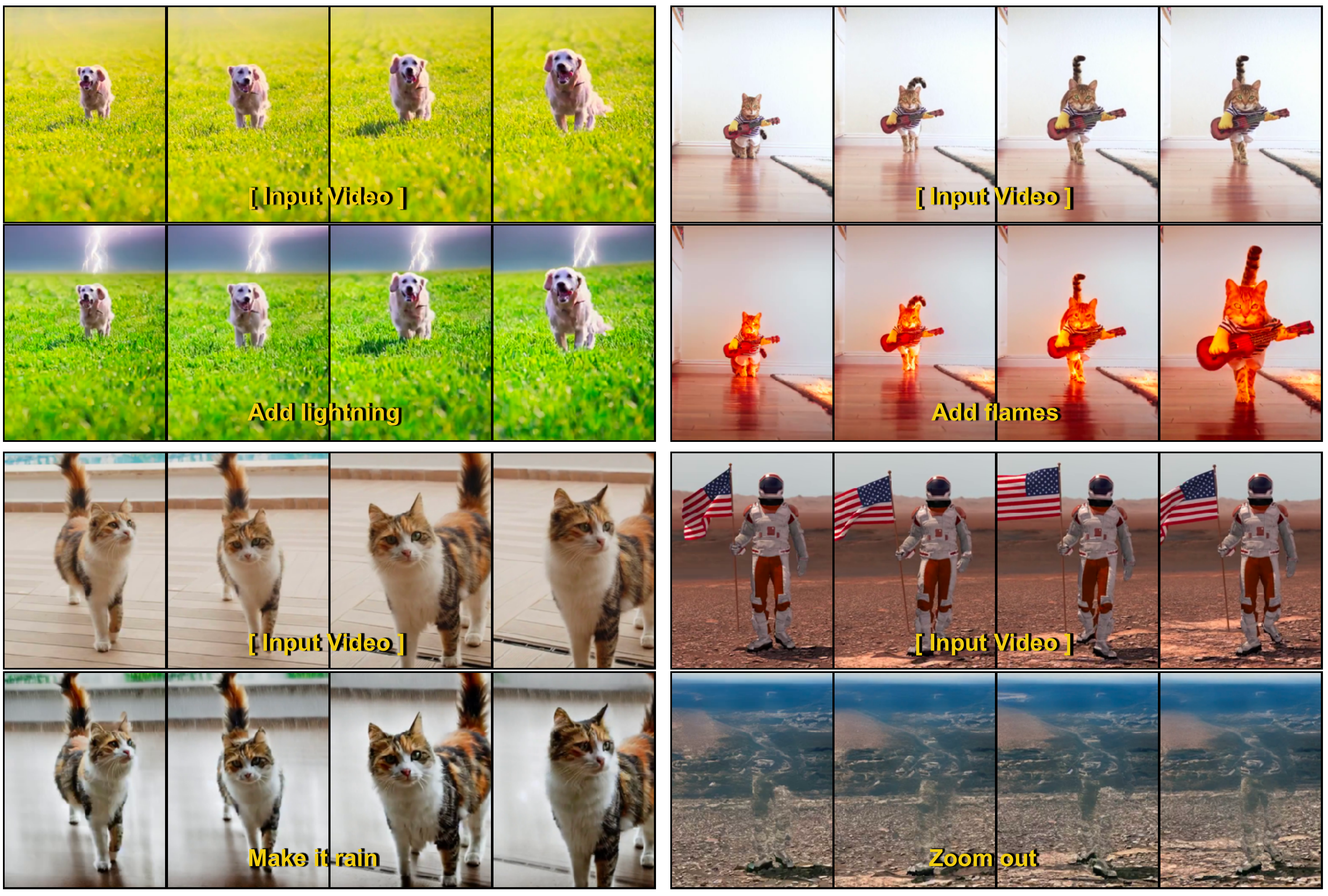}
\end{center}
\vspace{-5mm}
\caption{\textbf{Limitations of \ourmodel.} Our model cannot accurately render dynamic visual effects, such as lightning, flames, or rain.
} 
\vspace{-1mm}
\label{fig_appendix_limitation}
\end{figure*}

Finally, Figure \ref{fig_appendix_limitation} demonstrates some limitations we point out in section \ref{sec_limitations}. Since the model is never trained on video data, it does learn to generate concepts containing motion such as raining, lightning, or flames. \ourmodell also inherits the limit of the image editing model, where it is not able to follow the instructions that involve camera motion, such as zoom in or zoom out.